\documentclass{article}

\usepackage{microtype}
\usepackage{graphicx}
\usepackage{booktabs} 

\usepackage{subcaption}
\usepackage{tablefootnote}
\usepackage{multicol, multirow}
\usepackage{tabularx}
\usepackage{tikz} 

\definecolor{blue_colorblind}{RGB}{0, 90, 181}
\definecolor{red_colorblind}{RGB}{220, 50, 32}
\definecolor{violet_colorblind}{RGB}{117, 69, 102}

\definecolor{11}{RGB}{220, 38, 127}
\definecolor{21}{RGB}{100, 143, 255}
\definecolor{31}{RGB}{100, 143, 255}
\definecolor{41}{RGB}{255, 176, 0}
\definecolor{51}{RGB}{255, 176, 0}
\definecolor{61}{RGB}{220, 38, 127}

\definecolor{12}{RGB}{207, 133, 95}
\definecolor{22}{RGB}{160, 90, 191}
\definecolor{32}{RGB}{160, 90, 191}
\definecolor{42}{RGB}{237, 107, 63}
\definecolor{52}{RGB}{231, 84, 84}
\definecolor{62}{RGB}{191, 119, 127}

\definecolor{13}{RGB}{204, 100, 116}
\definecolor{23}{RGB}{175,	104,	159 }
\definecolor{33}{RGB}{183,	111,	143 } 
\definecolor{43}{RGB}{219,	108,	89 } 
\definecolor{53}{RGB}{211,	119,	95 } 
\definecolor{63}{RGB}{199,	102,	123 }

\definecolor{14}{RGB}{203	110	112
}
\definecolor{24}{RGB}{191	106	133
}
\definecolor{34}{RGB}{189	102	137
}
\definecolor{44}{RGB}{207	109	105
}
\definecolor{54}{RGB}{207	103	109
}
\definecolor{64}{RGB}{196	107	123
}

\usepackage{hyperref}

\usepackage[accepted]{icml2023}

\usepackage{amsmath}
\usepackage{amssymb}
\usepackage{mathtools}
\usepackage{amsthm}

\usepackage[capitalize,noabbrev]{cleveref}

\theoremstyle{plain}
\newtheorem{theorem}{Theorem}[section]
\newtheorem{proposition}[theorem]{Proposition}
\newtheorem{lemma}[theorem]{Lemma}

\theoremstyle{definition}

\theoremstyle{remark}

\usepackage[textsize=tiny]{todonotes}

\icmltitlerunning{Revisiting Over-smoothing and Over-squashing Using Ollivier-Ricci Curvature}

\usepackage{amsmath,amsfonts,bm}

\def\eqref#1{equation~\ref{#1}}
\def\Eqref#1{Equation~(\ref{#1})}

\def\1{\bm{1}}

\def\vx{{\bm{x}}}
\def\vy{{\bm{y}}}

\def\mH{{\bm{H}}}

\def\mW{{\bm{W}}}
\def\mX{{\bm{X}}}

\DeclareMathAlphabet{\mathsfit}{\encodingdefault}{\sfdefault}{m}{sl}
\SetMathAlphabet{\mathsfit}{bold}{\encodingdefault}{\sfdefault}{bx}{n}

\def\gE{{\mathcal{E}}}

\def\gG{{\mathcal{G}}}

\def\gV{{\mathcal{V}}}

\newcommand{\R}{\mathbb{R}}

\begin{document}

\twocolumn[
\icmltitle{Revisiting Over-smoothing and Over-squashing Using Ollivier-Ricci Curvature}



\icmlsetsymbol{equal}{*}

\begin{icmlauthorlist}
\icmlauthor{Khang Nguyen}{fsoft,hcmus}
\icmlauthor{Hieu Nong}{fsoft}
\icmlauthor{Vinh Nguyen}{fsoft}
\icmlauthor{Nhat Ho}{utaustin}
\icmlauthor{Stanley Osher}{ucla}
\icmlauthor{Tan Nguyen}{nus}
\end{icmlauthorlist}

\icmlaffiliation{fsoft}{FPT Software AI Center,  Vietnam}
\icmlaffiliation{ucla}{Department of Mathematics, University of California, Los Angeles, USA}
\icmlaffiliation{utaustin}{Department of Statistics and Data Sciences, University of Texas at Austin, USA}
\icmlaffiliation{hcmus}{Faculty of Mathematics and Computer Science, University of Science, Vietnam National University Ho Chi Minh City, Vietnam}
\icmlaffiliation{nus}{Department of Mathematics, National University of Singapore, Singapore}


\icmlcorrespondingauthor{Khang Nguyen}{khang.nguyenhoang.vn@gmail.com}
\icmlcorrespondingauthor{Tan Nguyen}{tanmnguyen89@gmail.com}

\icmlkeywords{Machine Learning, ICML}

\vskip 0.3in
]



\printAffiliationsAndNotice{}  

\begin{abstract}
Graph Neural Networks (GNNs) had been demonstrated to be inherently susceptible to the problems of over-smoothing and over-squashing. These issues prohibit the ability of GNNs to model complex graph interactions by limiting their effectiveness in taking into account distant information. Our study reveals the key connection between the local graph geometry and the occurrence of both of these issues, thereby providing a unified framework for studying them at a local scale using the Ollivier-Ricci curvature. Specifically, we demonstrate that over-smoothing is linked to positive graph curvature while over-squashing is linked to negative graph curvature. Based on our theory, we propose the Batch Ollivier-Ricci Flow, a novel rewiring algorithm capable of simultaneously addressing both over-smoothing and over-squashing.
\end{abstract}

\section{Introduction}
\label{sec:introduction}
A collection of entities with a set of relations between them is among the simplest, and yet most general, types of structure. It came as no surprise that numerous real world data are naturally represented by graphs~\cite{harary1967graph,  hsu2008graph, estrada2013}, motivating many recent advancements in the study of Graph Neural Networks (GNNs). This has lead to a wide range of successful applications, including physical modeling~\cite{battaglia2016interaction,kipf2018, sanchez_gonzalez_2018}, chemical and biological inference~\cite{duvenaud2015convolutional,gilmer2017}, recommender systems~\cite{berg2017,ying2018graph, fan2019, wu2019}, generative models~\cite{li2018learning, bojchevski2018netgan, shi2020graphaf}, financial prediction~\cite{chen2019stock, matsunaga2019exploring, yang2019financial}, and knowledge graphs \cite{Shang2019, zhang2019}.

{\color{black} Despite their success, popular GNN designs suffer from two notable setbacks that hamper their performance in practical applications that require long-range interactions.} {The first} common problem encountered by GNNs is known as over-smoothing \cite{li2018}. {Over-smoothing} occurs when node features quickly converge to each other and become indistinguishable as the number of layers increases. This issue {puts a limit on the depth of a GNN}, prohibiting {the model's capability of capturing} complex relationships {in the data}. Another plight plaguing GNNs is known as over-squashing \cite{alon2021}. This {phenomenon} happens when the number of nodes {within} the receptive field {of} a particular node grows exponentially with the number of layers, leading to the squashing of exponentially-growing amount of information into {fixed-size} node features. Such over-squashing {hinders the ability} of GNNs to {effectively process} distant information and {capture long-range dependencies between nodes in the graph, especially in the case of deep GNNs that require many layers}.

Together, over-smoothing and over-squashing impair the performance of modern GNNs, impeding their application to many important settings that involve very large graphs. \cite{cai2020,alon2021}. Understanding and alleviating either of these problems has been the main focus in many recent studies of GNNs \cite{oono2019, cai2020, zhao2020, karhadkar2022}. {\color{black} Notably, using a notion of graph curvature, \citet{topping2022} suggested that over-squashing behaviors could arise from local graph structures. This pioneering idea implies that local graph properties can be used to study and improve GNN performance.} It has been recently noted that over-smoothing and over-squashing are somewhat related problems \cite{karhadkar2022}. Nevertheless, to the best of our knowledge, there has been no work in the literature that offers a common framework to understand these problems. Such a unified approach presents a potentially crucial theoretical contribution to our understanding of the over-smoothing and over-squashing issues. It enables the development of novel architectures and methods that can effectively learn complex and long-range graph interactions, thereby broadening the applications of GNNs on practical tasks. 

\noindent
 \textbf{Contribution.}
We present a unified theoretical framework to study both the over-smoothing and over-squashing phenomena in GNNs at {the} local level using the Ollivier-Ricci curvature \cite{ollivier}, an inherent local geometric property of graphs. Our key contributions {are three-fold}:

\begin{enumerate} 
    \item We {prove} that very positively curved edges cause node representations to become similar, thereby establishing a link between the over-smoothing issue and high edge curvature. 
    \item We {prove} that low curvature value characterizes graph bottlenecks and demonstrate a connection between the over-squashing issue and negative edge curvature. 
    \item Based on our {theoretical results}, we propose {the} Batch Ollivier-Ricci Flow (BORF), a {novel} curvature-based rewiring method designed to mitigate the over-smoothing and over-squashing issues {simultaneously}.
\end{enumerate}

\vspace{0.5 em}
\noindent
\textbf{Organization.} We structure this paper as follows. First, we discuss related works in Section~\ref{sec:related_work}. In Section~\ref{sec:preliminaries}, we give a brief summary of the relevant backgrounds in the study of GNNs and provide a concise formulation for the Ollivier-Ricci curvature on graphs. In Section~\ref{sec:theoretical_analysis}, we present our central analysis showing positive graph curvature is associated with over-smoothing, while negative graph curvature is associated with over-squashing. In Section \ref{sec:borf}, we introduce the novel graph rewiring method BORF, {which} modifies the local graph geometry to suppress over-smoothing and {\color{black} alleviate} over-squashing inducing connections. We empirically demonstrate the {superior performance} of our method compared to other {state-of-the-art} rewiring methods in Section \ref{sec:experiments}. The paper ends with concluding remarks in Section~\ref{sec:conclusion}. Technical proofs and other additional materials are provided in the Appendix.

\vspace{0.5 em}
\noindent
\textbf{Notation.}
We denote scalars by lower- or upper-case letters and vectors and matrices by lower- and upper-case boldface letters, respectively. We use $\gG = (\gV,\gE)$ to denote a simple, connected graph $\gG$ with vertex set $\gV$ and edge set $\gE$. Graph vertices are also referred to as nodes, and the characters $u, v, w, p, q$ are reserved for representing them. We write $u \sim v$ if $(u,v) \in \gE$. The shortest path distance between two vertices $u,v$ is denoted by $d(u,v)$. We let $\mathcal{N}_u = \{p \in \gV \mid p \sim u\}$ be the $1$-hop neighborhood and $\Tilde{\mathcal{N}}_u = \mathcal{N}_u \cup \{u\}$ be the extended neighborhood of $u$. {\color{black} The characters $n,m$ are used to denote the degrees of the vertices $u, v$}.


\section{Related Work}\label{sec:related_work}
{\bf Over-smoothing:} First recognised by \cite{li2018}, who observed that GCN with non-linearity removed induces a smoothing effect on data features, over-smoothing has been one of the focal considerations in the study of GNNs. A dynamical system approach was used by \cite{oono2019} to show that under considerable assumptions, even GCN with ReLU can not escape this plight. Follow-up work by \cite{cai2020}  generalized and improved this approach. 
Designing ways to alleviate or purposefully avoid the problem is a lively research area \cite{luanzhao2019, zhao2020, rusch2022}. Notably, randomly removing edges from the base graph consistently improves GNN performance \cite{rong2020dropedge}.

\noindent
{\bf Over-squashing:} The inability of GNNs to effectively take into account distant information has long been observed \cite{xu2018}. \citet{alon2021} showed that this phenomenon cound be explained by the existence of local bottlenecks in the graph structure. {\color{black} It was shown by \cite{topping2022} that graph curvature provided an insightful way to study and address the over-squashing problem.} Methods have been designed to tackle over-squashing, including those by \cite{banerjee2022} and \cite{karhadkar2022}.

\noindent
{\bf Graph curvature:} Efforts have been made to extend the geometric notion of curvature to settings other than smooth manifolds, including on graphs \cite{bakryemery1985,Forman2003}. Among these, the Ollivier's Ricci curvature \cite{ollivier} is arguably the superior attempt due to its proven compatibility with the classical notion of curvature in differential geometry \cite{ollivier, hoorn2020}. Graph curvature has been utilised in the study of complex networks \cite{ni2015, sia2019}, and a number of works have experimented with its use in GNNs \cite{topping2022,Bober2022}.

\section{Preliminaries}
\label{sec:preliminaries}
We begin by summarizing the relevant backgrounds on message passing neural networks and the over-smoothing and over-squashing issues of GNNs. We also provide a concise formulation for the Ollivier-Ricci curvature on graphs.

\subsection{Message Passing Neural Networks}  
\label{subsec:mpnn}
Message passing neural networks (MPNNs) \cite{gilmer2017} are a unified framework for a broad range of graph neural networks. It encompasses virtually every popular GNN design to date, including graph convolutional network (GCN) \cite{kipf2016}, GraphSAGE \cite{hamilton2017}, graph attention network (GAT) \cite{velickovic2018}, graph isomorphism network (GIN) \cite{xu2019}, etc. The key idea behind MPNNs is that by aggregating information from local neighborhoods, a neural network can effectively use both node feature data and the graph topology to learn relevant information. \par 
Let $\mX \in \R^{|\gV| \times d}$ be the node feature matrix of a graph $\gG$, where $d$ is the number of feature channels. Let $\mX^k$ be the node feature matrix at layer $k$, with the convention that $\mX^0 = \mX$. The features of node $u$ at layer $k$ is denoted by $\mX^k_u$, and is exactly the transpose of the $u$-th row of $\mX^k$. A general formulation for an MPNN can be given by
\begin{equation} \label{eq:general_mpnn}
    \mX^{k+1}_u = \phi_k \left(\bigoplus_{p \in \Tilde{\mathcal{N}}_u} \psi_k(\mX^k_p) \right),
\end{equation}
where $\psi_k$ is a message function, $\bigoplus$ is an aggregating function, and $\phi_k$ is an update function. Table \ref{table:popular_gnn} summarizes the choice for $\psi_k, \phi_k$, and $\bigoplus$ in four popular GNN {architectures}.
We give further discussion on how \Eqref{eq:general_mpnn} accommodates different designs of GNNs in Appendix \ref{sec:mpnn}. From now on, we will use MPNN and GNN interchangeably.

\begin{table}[t!]
\caption{Popular GNNs are instances of \Eqref{eq:general_mpnn}: GCN \cite{kipf2016}, GraphSAGE \cite{hamilton2017}, GAT \cite{velickovic2018}, and GIN \cite{xu2019}.}
\label{table:popular_gnn}
\vskip 0.1in
\begin{center}
\resizebox{\linewidth}{!}{
\begin{tabular}{cccc}
\multicolumn{1}{c}{\bf GNN}  &\multicolumn{1}{c}{$\psi_k$} & \multicolumn{1}{c}{ \centering$\phi_k$} &\multicolumn{1}{c}{$\bigoplus$}
\\ 
\midrule
GCN \tablefootnote{If the symmetrically normalized Laplacian is replaced by the normalized Laplacian. See Appendix \ref{sec:mpnn}.} 
& linear & activation   & mean \\
GraphSAGE \tablefootnote{Mean aggregator variant.}
& linear  & activation & mean\\
GAT 
& linear  & activation & weighted mean \\
GIN \tablefootnote{GIN-$0$ variant.} 
& identity & MLP & sum
\end{tabular}
}
\end{center}
\vspace{-1em}
\end{table}

Traditionally, GNNs are designed to operate directly on the input graphs. In many cases, this leads to significant downsides due to possible undesirable characteristics of the dataset.  Hence, it has been proposed that by conducting the learning process on a modified version of the input graphs, we can improve upon the scale and performance of graph models \cite{hamilton2017, digl}. One such approach is known as graph rewiring, which involves modifying the set of edges $\gE$ within a graph as a preprocessing step. We give a brief overview of two novel rewiring algorithms, SDRF \cite{topping2022} and FoSR \cite{karhadkar2022}, along with a comparison between them and our proposed method in Section \ref{sec:borf}.

\subsection{The Over-smoothing and Over-squashing Issues of GNNs}
\label{subsec:over_smoothing_squashing}
Over-smoothing has generally been described as the phenomenon where the feature representation of every node becomes similar to each other as the number of layers in a GNN increases \cite{li2018}. If over-smoothing occurs, for every two neighbor nodes $u,v$, it must happen that
\begin{equation}
    \label{eq:local_over-smoothing}
    \left|\mX_u^{k} - \mX_v^{k} \right| \to 0 \text{ as } k \to \infty.
\end{equation}
\Eqref{eq:local_over-smoothing} can be thought of as the local smoothing behavior, observed in the neighborhood of two nodes $u \sim v$.

A global formulation for feature representation similarity is obtained by summing up terms of the form $\left|\mX_u^{k} - \mX_v^{k} \right|$ for all neighbors $u,v$. Formally, we obtain a formulation for the global over-smoothing issue based on local observations 
\begin{equation}
    \label{eq:over-smoothing}
    \sum_{(u,v) \in \gE}\left|\mX_u^{k} - \mX_v^{k} \right| \to 0 \text{ as } k \to \infty.
\end{equation}
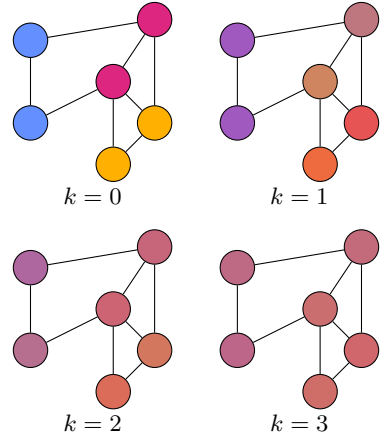
\begin{figure}[t!]
\vskip 0.1in
    \centering
        \begin{tikzpicture}[scale=0.55, uv/.style = {draw, circle, red, minimum size=2em}, neighbor/.style = {draw, circle, minimum size=1.3em}] 
        \node[neighbor , at={(0,0)}, fill = 11] (1) {}; 
        \node[neighbor , at={(-2,1)}, fill = 21] (2) {}; 
        \node[neighbor ,at={(-2,-1)}, fill = 31] (3) {}; 
        \node[neighbor ,at={(0,-2)}, fill = 41] (4) {};
        \node[neighbor,at={(1,-1)}, fill = 51] (5) {};
        \node[neighbor ,at={(1,1.5)}, fill = 61] (6) {};
        \draw (1) -- (3);
        \draw (2) -- (3);
        \draw (2) -- (6);
        \draw (1) -- (6);
        \draw (6) -- (5);
        \draw (5) -- (1);
        \draw (4) -- (1);
        \draw (5) -- (4);
        
        \node[neighbor , at={(5,0)}, fill = 12] (11) {}; 
        \node[neighbor , at={(3,1)}, fill = 22] (12) {}; 
        \node[neighbor ,at={(3,-1)}, fill = 32] (13) {}; 
        \node[neighbor ,at={(5,-2)}, fill = 42] (14) {};
        \node[neighbor,at={(6,-1)}, fill = 52] (15) {};
        \node[neighbor ,at={(6,1.5)}, fill = 62] (16) {};
        \draw (11) -- (13);
        \draw (12) -- (13);
        \draw (12) -- (16);
        \draw (11) -- (16);
        \draw (16) -- (15);
        \draw (15) -- (11);
        \draw (14) -- (11);
        \draw (15) -- (14);
        
        \node[neighbor , at={(0,-5.5)}, fill = 13] (21) {}; 
        \node[neighbor , at={(-2,-4.5)}, fill = 23] (22) {}; 
        \node[neighbor ,at={(-2,-6.5)}, fill = 33] (23) {}; 
        \node[neighbor ,at={(0,-7.5)}, fill = 43] (24) {};
        \node[neighbor,at={(1,-6.5)}, fill = 53] (25) {};
        \node[neighbor ,at={(1, -4)}, fill = 63] (26) {};
        \draw (21) -- (23);
        \draw (22) -- (23);
        \draw (22) -- (26);
        \draw (21) -- (26);
        \draw (26) -- (25);
        \draw (25) -- (21);
        \draw (24) -- (21);
        \draw (25) -- (24);
        
        \node[neighbor , at={(5,-5.5)}, fill = 14] (31) {}; 
        \node[neighbor , at={(3,-4.5)}, fill = 24] (32) {}; 
        \node[neighbor ,at={(3,-6.5)}, fill = 34] (33) {}; 
        \node[neighbor ,at={(5,-7.5)}, fill = 44] (34) {};
        \node[neighbor,at={(6,-6.5)}, fill = 54] (35) {};
        \node[neighbor ,at={(6, -4)}, fill = 64] (36) {};
        \draw (31) -- (33);
        \draw (32) -- (33);
        \draw (32) -- (36);
        \draw (31) -- (36);
        \draw (36) -- (35);
        \draw (35) -- (31);
        \draw (34) -- (31);
        \draw (35) -- (34);
        
        \node at (-0.5,-2.75) {\small $k = 0$};
        \node at (4.5,-2.75) {\small $k = 1$};
        \node at (-0.5,-8.25) {\small $k = 2$};
        \node at (4.5,-8.25) {\small $k = 3$};
    \end{tikzpicture} 
    \vspace{-0.5em}
    \caption{Over-smoothing induced by the averaging operation.}
    \label{fig:oversmoothing_illustration}
\vskip -0.1in
\end{figure}
 That is, if the term $\sum_{(u,v) \in \gE}\left|\mX_u^{k} - \mX_v^{k} \right|$ converges to zero, we say that the model experiences over-smoothing. This formulation is similar to the definition based on the node-wise Dirichlet energy utilised in \cite{rusch2022}. The Dirichlet energy was first proposed as a viable way to measure the over-smoothing issue by \cite{cai2020}. Figure \ref{fig:oversmoothing_illustration} visualizes the over-smoothing behavior of a simple graph containing $6$ nodes from $3$ classes with different RGB color features. At the start, the nodes can easily be divided into $3$ classes. When the mean operator is applied repeatedly for $k$ times across $k$ GNN layers, those nodes rapidly converge to having similar colors. At the final step $k=3$, the nodes in the GNN have become virtually indistinguishable, suggesting that they have experienced over-smoothing.

\begin{figure}[t]
\vskip 0.05in
    \centering
    \begin{tikzpicture}[scale = 0.6, uv/.style = {draw, circle, minimum size=1em}, neighbor/.style = {draw, circle, minimum size=1em}] 
        \node[uv , at={(0,0)}] (1) {}; 
        \node[uv , at={(1,0)}] (2) {}; 
        \node[neighbor, at={(-1, 1.875)}] (3) {};
        \node[neighbor, at={(-1, 0)}] (4) {};
        \node[neighbor, at={(-1, -1.875)}] (5) {};
        \node[neighbor, at={(-2.5, 1.875)}] (6) {};
        \node[neighbor, at={(-2.5, 0)}] (7) {};
        \node[neighbor, at={(-2.5,-1.875)}] (8) {};
        \node[neighbor, at={(-2.5 , 2.5)}] (9) {};
        \node[neighbor, at={(-2.5, 1.25)}] (10) {};
        \node[neighbor, at={(-2.5, 0.625)}] (11) {};
        \node[neighbor, at={(-2.5 ,-0.625)}] (12) {};
        \node[neighbor, at={(-2.5,-1.25)}] (13) {};
        \node[neighbor, at={(-2.5,-2.5)}] (14) {};
        \node[neighbor, at={(2, 1.875)}] (15) {};
        \node[neighbor, at={(2, 0)}] (16) {};
        \node[neighbor, at={(2, -1.875)}] (17) {};
        \node[neighbor, at={(3.5, 1.875)}] (18) {};
        \node[neighbor, at={(3.5, 0)}] (19) {};
        \node[neighbor, at={(3.5,-1.875)}] (20) {};
        \node[neighbor, at={(3.5 , 2.5)}] (21) {};
        \node[neighbor, at={(3.5, 1.25)}] (22) {};
        \node[neighbor, at={(3.5, 0.625)}] (23) {};
        \node[neighbor, at={(3.5 ,-0.625)}] (24) {};
        \node[neighbor, at={(3.5,-1.25)}] (25) {};
        \node[neighbor, at={(3.5,-2.5)}] (26) {};
        
        \draw (1) -- (3);
        \draw (1) -- (4);
        \draw (1) -- (5);
        \draw (3) -- (6);
        \draw (4) -- (7);
        \draw (5) -- (8);
        \draw (3) -- (9);
        \draw (3) -- (10);
        \draw (4) -- (11);
        \draw (4) -- (12);
        \draw (5) -- (13);
        \draw (5) -- (14);
        \draw (2) -- (15);
        \draw (2) -- (16);
        \draw (2) -- (17);
        \draw (21) -- (15);
        \draw (22) -- (15);
        \draw (18) -- (15);
        \draw (23) -- (16);
        \draw (19) -- (16);
        \draw (24) -- (16);
        \draw (20) -- (17);
        \draw (25) -- (17);
        \draw (26) -- (17);
        \draw[red_colorblind, line width=2pt] (1) -- (2);
        
        \node[uv , at={(7,0)}] (31) {}; 
        \node[uv , at={(8,0)}] (32) {}; 
        \node[neighbor, at={(4.5, 0)}] (33) {};
        \node[neighbor, at={(5, 2)}] (34) {};
        \node[neighbor, at={(5.25, -1.75)}] (35) {};
        \node[neighbor, at={(5.5, 1)}] (36) {};
        \node[neighbor, at={(6, -0.625)}] (37) {};
        \node[neighbor, at={(9.75, 1.75)}] (38) {};
        \node[neighbor, at={(10 , 0.75)}] (39) {};
        \node[neighbor, at={(9, 1)}] (310) {};
        \node[neighbor, at={(9, -0.75)}] (311) {};
        \node[neighbor, at={(9.25 , 2.375)}] (312) {};
        \node[neighbor, at={(10.5,-0.625)}] (313) {};
        \node[neighbor, at={(9.5,-2)}] (314) {};
        
        \draw (31) -- (36);
        \draw (31) -- (37);
        \draw (36) -- (37);
        \draw (35) -- (37);
        \draw (34) -- (36);
        \draw (34) -- (33);
        \draw (33) -- (35);
        \draw (33) -- (37);
        \draw (32) -- (310);
        \draw (32) -- (311);
        \draw (310) -- (312);
        \draw (310) -- (38);
        \draw (38) -- (312);
        \draw (38) -- (39);
        \draw (311) -- (313);
        \draw (311) -- (314);
        \draw (313) -- (314);
        \draw (39) -- (313);
        \draw (39) -- (310);
        \draw[red_colorblind, line width = 2pt] (31) -- (32);
\end{tikzpicture} 
    \vspace{-0.5em}
    \caption{Bottlenecks inhibit the message passing capability of MPNNs. These bottlenecks are highlighted by bold red lines.}
    \label{fig:oversquashing_illustration}
\vskip -0.1in
\end{figure}
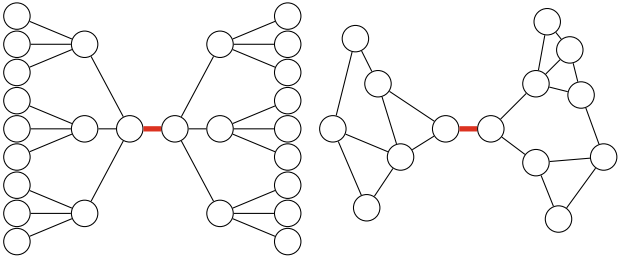

On the other hand, over-squashing is an inherent pitfall of GNNs that occurs when bottlenecks in the graph structure impede the graph's ability to propagate information among its vertices. We observe from \Eqref{eq:general_mpnn} that messages can only be transmitted by a distance of $1$ at each layer. Hence, two nodes of distance $K$ receive information from each other if and only if the GNN has at least $K$ layers. For any given node $u$, the set of nodes whose messages can reach $u$ is called the receptive field of $u$. As the number of layers increases, the size of each node's receptive field increases exponentially \cite{chen2018stochastic}. This causes messages between exponentially-growing number of distant vertices to be squashed into fixed size vectors, limiting {the model's} ability to capture long range dependencies \cite{alon2021}. As illustrated in Figure \ref{fig:oversquashing_illustration}, graph bottlenecks contribute to this problem by enforcing the maximal rate of expansion to the receptive field, while providing minimal connection between either sides of the bottleneck. Thus, a graph containing many bottlenecks causes GNNs to suffer from over-squashing.




\subsection{Ollivier-Ricci Curvature on Graph}
\label{subsec:ollivier}
The Ricci curvature is a geometric object ubiquitous in the field of differential geometry. 
At a local neighborhood of a Riemannian manifold, the Ricci curvature of the space characterizes the average geodesic dispersion, i.e., whether straight paths in a given direction of nearby points have the tendency to remain parallel (zero curvature), converge (positive curvature), or  diverge (negative curvature). Crucially, the definition of the Ricci curvature depends on the ability to specify directions, or more precisely, tangent vectors, within the space considered.

To circumvent the lack of a tangent structure on graphs, the Ollivier-Ricci curvature \cite{ollivier} considers random walks from nearby points. We define a random walk $\mu$ on a graph $\gG$ as a family of probability measure $\mu_u(\cdot)$ on the vertex set $\gV$ for all $u \in \gV$. For a vertex $p \in \gV$, it is intuitive to think of $\mu_u(p)$ as the probability that a random walker starting from $u$ will end up at $p$ after some number of steps. Then, for any $u,v \in \gV$, we can consider the $L^1$ Wasserstein transport distance $W_1(\mu_u, \mu_v)$ given by 
\begin{equation*}
    W_1(\mu_u, \mu_v) = \inf_{\pi \in \Pi(\mu_u, \mu_v)} \left(\sum_{(p,q) \in \gV^2} \pi(p,q) d(p,q) \right),
\end{equation*}
where $\Pi(\mu_u, \mu_v)$ is the family of joint probability distributions of $\mu_u$ and $\mu_v$. This measures\footnote{More precisely, $W_1(\mu_u,\mu_v)$ measures the minimal distance one must move the random walk from $u$ in order to obtain the random walk from $v$.} the minimal distance that random walkers from $u$ must travel to meet the random walkers from $v$. 
 The Ollivier-Ricci curvature $\kappa(u,v)$ is then defined based on the ratio between the random walker distance $W_1(\mu_u, \mu_v)$ and the original distance $d(u,v)$ 
\begin{equation}
    \label{eq:curvature_def}
    \kappa(u,v) = 1 - \frac{W_1(\mu_u, \mu_v)}{d(u,v)}.
\end{equation}
Such a definition captures the behavior that $\kappa(u,v) = 0$ if the random walkers tend to stay at equal distance, $\kappa(u,v) < 0$ if they tend to diverge, and $\kappa(u,v) > 0$ if they tend to converge.

Since curvature is intrinsically a local concept, it makes sense to only examine small neighborhoods when defining any curvature notion. On Riemannian manifolds, various definitions of curvature are constructed using differentials and derivatives on arbitrarily small neighborhoods. On a graph, the smallest neighborhood has radius $1$, and so it is natural to consider the uniform $1$-step random walk $\mu$ given by
\begin{equation*}
    \mu_u(p) = \begin{cases}
    \frac{1}{\operatorname{deg} u} & \text{ if $p \sim u$,}\\
    0 & \text{ otherwise.}
    \end{cases}
\end{equation*}
Hence, the Ollivier-Ricci curvature on graphs $\kappa$ is defined by \Eqref{eq:curvature_def}, where $W_1(\mu_u, \mu_v)$ is the optimal value of the objective function in the linear optimization problem 
\begin{align}
\text{minimize} &\sum_{p \in \mathcal{N}_u} \sum_{q \in \mathcal{N}_v} d(p,q) \pi(p,q),& \nonumber
\\
\text{subject to} &\sum_{p \in \mathcal{N}_u}\pi(p,q) = \tfrac{1}{\deg v}, \label{eq:W1distance} \\
&\sum_{q \in \mathcal{N}_v}\pi(p,q) = \tfrac{1}{\deg u}.\nonumber
\end{align}

{Our analysis in Section \ref{sec:theoretical_analysis} is based on this specific formulation of the Ollivier-Ricci curvature. }

\section{Analysis Based on Graph Curvature}
\label{sec:theoretical_analysis}

Throughout this section, we assume $u \sim v \in \gV$ are neighboring vertices with $\operatorname{deg} u = n, \operatorname{deg} v = m$, and $n \geq m$. {We note that $u \sim v$ implies $0 \leq d(p,q) \leq 3$ for all neighbors $p,q$ of $u,v$, and so $0 \leq W_1(\mu_u, \mu_v) \leq 3$. From \Eqref{eq:curvature_def}, the following bound holds $$-2 \leq \kappa(u,v) \leq 1.$$} Hence, a curvature value close to $1$ is considered very positive, while a value close to $-2$ is considered very negative. 

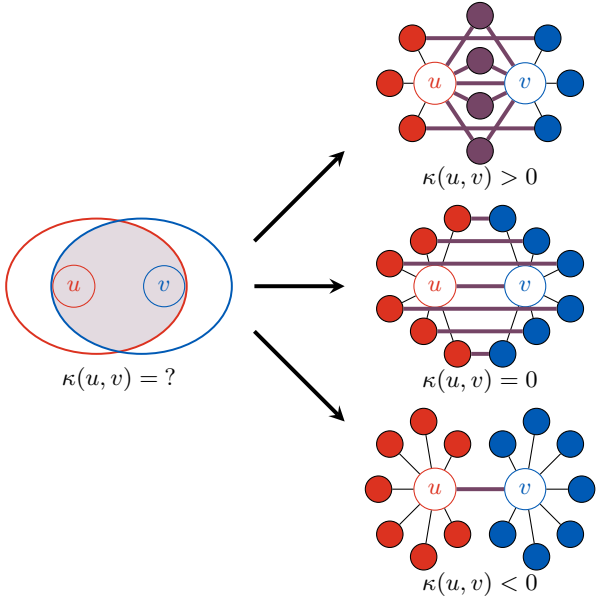
\begin{figure}[t!]
\vskip 0.1in
    \centering
    \begin{tikzpicture}[scale = 0.6, u/.style = {draw, circle, red_colorblind, minimum size= 1em},
    v/.style = {draw, circle, blue_colorblind, minimum size= 1em},
    neighboru/.style = {draw, circle, fill = red_colorblind, minimum size=1em}, neighborv/.style = {draw, circle, fill = blue_colorblind, minimum size=1em},
    neighboruv/.style = {draw, circle, fill = violet_colorblind, minimum size=1em}] 

        \def\firstellipse{(-7.5,0) ellipse (2 and 1.5)}
        \def\secondellipse{(-6.5,0) ellipse (2 and 1.5)}
    
    
        \begin{scope}
            \clip \firstellipse;
            \fill[violet_colorblind!20] \secondellipse;
        \end{scope}
    
        \draw[red_colorblind, , line width = 0.75] \firstellipse;
        \draw[blue_colorblind, line width = 0.75] \secondellipse;
    
        \draw [-stealth, line width = 1.5](-4,0) -- (-2,0);
        \draw [-stealth, line width = 1.5](-4,1) -- (-2,3);
        \draw [-stealth, line width = 1.5](-4,-1) -- (-2,-3);
            
        \node[u , at={(-8,0)}] (1) {\small $u$}; 
        \node[v , at={(-6,0)}] (2) {\small $v$}; 
        
        \node at (-7,-2.05) {\small $\kappa(u,v) = \ ?$};
            
        \node[u , at={(0,4.5)}] (1) {\small $u$}; 
        \node[v , at={(2,4.5)}] (2) {\small $v$}; 
        \node[neighboruv , at={(1,6)}] (3) {}; 
        \node[neighboruv , at={(1,3)}] (4) {}; 
        \node[neighboruv , at={(1,5)}] (5) {}; 
        \node[neighboruv , at={(1,4)}] (6) {}; 
        \node[neighboru , at={(-0.5,5.5)}] (7) {}; 
        \node[neighboru , at={(-0.5,3.5)}] (8) {}; 
        \node[neighboru , at={(-1,4.5)}] (9) {}; 
        \node[neighborv , at={(2.5,5.5)}] (11) {}; 
        \node[neighborv , at={(2.5,3.5)}] (12) {}; 
        \node[neighborv , at={(3,4.5)}] (13) {}; 
        
        \draw[violet_colorblind, line width = 1.5pt] (1) -- (3);
        \draw[violet_colorblind, line width = 1.5pt] (1) -- (4);
        \draw[violet_colorblind, line width = 1.5pt] (1) -- (5);
        \draw[violet_colorblind, line width = 1.5pt] (1) -- (6);
        \draw[violet_colorblind, line width = 1.5pt] (2) -- (3);
        \draw[violet_colorblind, line width = 1.5pt] (2) -- (4);
        \draw[violet_colorblind, line width = 1.5pt] (2) -- (5);
        \draw[violet_colorblind, line width = 1.5pt] (2) -- (6);
        \draw (1) -- (7);
        \draw (1) -- (8);
        \draw (1) -- (9);
        \draw (2) -- (11);
        \draw (2) -- (12);
        \draw (2) -- (13);
        \draw[violet_colorblind, line width = 1.5pt] (7) -- (11);
        \draw[violet_colorblind, line width = 1.5pt] (8) -- (12);
        \draw[violet_colorblind, line width = 1.5pt] (1) -- (2);
        
        \node at (1, 2.4) {\small $\kappa(u,v) > 0$};
            
        \node[u , at={(0,0)}] (11) {\small $u$}; 
        \node[v , at={(2,0)}] (12) {\small $v$}; 
        \node[neighborv , at={(2.25,1)}] (13) {}; 
        \node[neighborv , at={(2.25,-1)}] (14) {}; 
        \node[neighboru , at={(-0.25,1)}] (15) {}; 
        \node[neighboru , at={(-0.25,-1)}] (16) {}; 
        \node[neighboru , at={(0.5,1.5)}] (17) {}; 
        \node[neighboru , at={(0.5,-1.5)}] (18) {}; 
        \node[neighboru , at={(-1,0.5)}] (19) {}; 
        \node[neighboru , at={(-1,-0.5)}] (110) {};
        \node[neighborv , at={(1.5,1.5)}] (111) {}; 
        \node[neighborv , at={(1.5,-1.5)}] (112) {}; 
        \node[neighborv , at={(3,0.5)}] (113) {}; 
        \node[neighborv , at={(3,-0.5)}] (114) {}; 
        
        \draw (12) -- (13);
        \draw (12) -- (14);
        \draw (11) -- (15);
        \draw (11) -- (16);
        \draw (11) -- (17);
        \draw (11) -- (18);
        \draw (11) -- (19);
        \draw (11) -- (110);
        \draw (12) -- (111);
        \draw (12) -- (112);
        \draw (12) -- (113);
        \draw (12) -- (114);
        \draw[violet_colorblind, line width = 1.5pt] (17) -- (111);
        \draw[violet_colorblind, line width = 1.5pt] (15) -- (13);
        \draw[violet_colorblind, line width = 1.5pt] (16) -- (14);
        \draw[violet_colorblind, line width = 1.5pt] (18) -- (112);
        \draw[violet_colorblind, line width = 1.5pt] (11) -- (12);
        \draw[violet_colorblind, line width = 1.5pt] (19) -- (113);
        \draw[violet_colorblind, line width = 1.5pt] (110) -- (114);
        
        \node at (1,-2.1) {\small $\kappa(u,v) = 0$};
        
        \node[u , at={(0,-4.5)}] (21) {\small $u$}; 
        \node[v , at={(2,-4.5)}] (22) {\small $v$}; 
        \node[neighborv , at={(2.25,-3)}] (23) {}; 
        \node[neighborv , at={(2.25,-6)}] (24) {}; 
        \node[neighboru , at={(-0.25,-3)}] (25) {}; 
        \node[neighboru , at={(-0.25,-6)}] (26) {}; 
        \node[neighboru , at={(0.5,-3.5)}] (27) {}; 
        \node[neighboru , at={(0.5,-5.5)}] (28) {}; 
        \node[neighboru , at={(-1,-3.5)}] (29) {}; 
        \node[neighboru , at={(-1,-5.5)}] (210) {};
        \node[neighboru , at={(-1.25,-4.5)}] (215) {}; 
        \node[neighborv , at={(1.5,-3.5)}] (211) {}; 
        \node[neighborv , at={(1.5,-5.5)}] (212) {}; 
        \node[neighborv , at={(3,-3.5)}] (213) {}; 
        \node[neighborv , at={(3,-5.5)}] (214) {}; 
        \node[neighborv , at={(3.25,-4.5)}] (216) {}; 
        
        \draw (22) -- (23);
        \draw (22) -- (24);
        \draw (21) -- (25);
        \draw (21) -- (26);
        \draw (21) -- (27);
        \draw (21) -- (28);
        \draw (21) -- (29);
        \draw (21) -- (210);
        \draw (21) -- (215);
        \draw (22) -- (211);
        \draw (22) -- (212);
        \draw (22) -- (213);
        \draw (22) -- (214);
        \draw (22) -- (216);
        \draw[violet_colorblind, line width = 1.5pt] (21) -- (22);
        
        \node at (1,-6.6) {\small $\kappa(u,v) < 0$};

    \end{tikzpicture} 
    \vspace{-0.8em}
    \caption{Different edge curvatures give rise to different local graph structures.}
    \label{fig:curvature_illustration}
\vskip -0.1in
\end{figure}

To motivate our findings, we remark that the curvature $\kappa(u,v)$ characterizes how well-connected the neighborhoods $\Tilde{\mathcal{N}}(u)$ and $\Tilde{\mathcal{N}}(v)$ are. Figure \ref{fig:curvature_illustration} illustrates how different local graph structures give rise to different graph curvature. Red and blue are used to color the neighborhoods $\Tilde{\mathcal{N}}_u \backslash \{v\}$ and $\Tilde{\mathcal{N}}_v \backslash \{u\}$. The color violet is used to signal shared vertices or edges connecting from one neighborhood to the other. If the neighborhoods mostly coincide then the transport cost is very low, leading to a positive curvature value. In this case, messages can be transmitted freely and easily between both neighborhoods. In contrast, if the neighborhoods only have minimal connections then the transport cost is high, leading to a negative curvature value. Each such connection will then act as a bottleneck, limiting the effectiveness of the message-passing mechanism. 

\subsection{Positive Graph Curvature and Over-smoothing}
We identify the key connection between positive graph curvature and the occurrence of the over-smoothing issue. 
\begin{lemma} \label{thm:positive_curvature}
The following inequality holds
\begin{equation*}
    \frac{|\mathcal{N}_u \cap \mathcal{N}_v |}{\max(m,n)} \geq \kappa(u,v).
\end{equation*}
\end{lemma}
Lemma \ref{thm:positive_curvature} says that the curvature $\kappa(u,v)$ is a lower bound for the proportion of shared neighbors between $u$ and $v$. A closer inspection of \Eqref{eq:general_mpnn} reveals a fundamental characteristic of GNNs: at the $k$-th layer, every node $p$ broadcasts an identical message $\psi_k(\mX_p^k)$ to each vertex $u$ in its $1$-hop neighborhood. These messages are then aggregated and used to update the features of $u$. If $\kappa(u,v)$ is very positive then the neighborhoods $\mathcal{N}_u$ and $\mathcal{N}_v$ mostly coincide. Hence, they incorporate roughly the same information, and their variance diminishes. This gives us significant insight into why over-smoothing happens, and is made precise by the following theorem.

\begin{theorem} \label{thm:onelayerGNN}
Consider the update rule given by \Eqref{eq:general_mpnn}. Suppose the edge curvature $\kappa(u,v) > 0$. For some $k$, assume the update function $\phi_k$ is $L$-Lipschitz, $\left|\mX^k_p\right| \leq C$ for all $p \in \mathcal{N}(u) \cup \mathcal{N}(v)$, and the message function $\psi_k$ is bounded, i.e. $|\psi_k(\vx)| \leq M |\vx|, \forall \vx$. There exists a positive function $h: (0,1) \to \R^{+}$ dependent on the constants $L, M, C, n$ satisfying
\begin{itemize}
    \item if $\bigoplus$ is the sum operation then $h$ is constant;
    \item if $\bigoplus$ is the mean operation then $h$ is decreasing;
\end{itemize}
such that
\begin{equation}
    \left |\mX^{k+1}_u - \mX^{k+1}_v \right |  \leq (1-\kappa(u,v)) h(\kappa(u,v)).
\end{equation}
In both cases, we clearly have 
\begin{equation}
    \lim_{x \to 1} (1-x) h(x)  = 0.
\end{equation}
\end{theorem}

This result applies to a wide range of GNNs, including those in Table \ref{table:popular_gnn} with the exception of GAT, due to the fact that GAT employs the attention mechanism to create a learnable weighted mean function as the aggregator. Nevertheless, if the variance between attention weights are low, we expect the general behavior to still hold true.

Theorem \ref{thm:onelayerGNN} conclusively shows that  very positively curved edges force local node features to become similar. {If the graph is positively curved everywhere or if it contains multiple subgraphs having this characteristic, we can expect that the node features will quickly converge to indistinguishable representations. In other words, positive edge curvature induces the mixing behavior observed by~\cite{li2018}, causing over-smoothing to occur at a faster rate. This suggests the occurrence of over-smoothing in shallow GNNs can be explained by an abundance of positively curved edges.}


Any global analysis of the issue based on local observations is hindered by the complexity in dealing with graph structures. Nevertheless, by restricting our attention to a more manageable class of graphs - the class of regular graphs, we obtain Proposition \ref{prop:pos_curved_regular_multilayer_GNN}. This serves to illustrate how positive local graph curvature can affect the long term global behavior of a typical GNN.

\begin{proposition} \label{prop:pos_curved_regular_multilayer_GNN}
Consider the update rule given by \Eqref{eq:general_mpnn}. Assume the graph is regular. Suppose there exists a constant $\delta> 0$ such that for all edges $(u,v) \in \gE$, the curvature is bounded by $\kappa(u,v) \geq \delta > 0$.  For all $k \geq 1$, assume the functions $\phi_k$ are $L$-Lipschitz, $\bigoplus$ is realised as the mean operation, $\left|\mX^0_p\right| \leq C$ for all $p \in \gV$, and the message functions $\psi_k$ are bounded linear operators, i.e. $|\psi_k(\vx)| \leq M |\vx|, \forall \vx$. The following inequality holds for $k \geq 1$ and any neighboring vertices $u \sim v$
\begin{equation} \label{eq:multilayerGNNbound}
    \left|\mX^{k}_u - \mX^{k}_v \right| \leq \frac{2}{3} C \left(\frac{3LM\lfloor(1-\delta) n \rfloor}{n+1}\right)^k.
\end{equation}
Furthermore, for any $u, v \in \gV$ that are not necessarily neighbors, the following inequality holds
\begin{equation} \label{eq:multilayerGNNbound2}
    \left|\mX^{k}_u - \mX^{k}_v \right| \leq \frac{2}{3} \left\lfloor \frac{2}{\delta} \right\rfloor  C \left(\frac{3LM\lfloor(1-\delta) n \rfloor}{n+1}\right)^k.
\end{equation}
\end{proposition}

The conclusion of Proposition \ref{prop:pos_curved_regular_multilayer_GNN} says that if every edge curvature in a regular graph $G$ is bounded from below by a sufficiently high constant $\delta$ then the difference between the features of any pair of neighboring nodes, or indeed, any pair of nodes at all, exponentially converges to $0$ in a typical GNN. This leads to the over-smoothing issue  formulated in Equation (\ref{eq:over-smoothing}) since for appropriate constants $C_1, C_2 > 0$, we have
\begin{equation*} 
    \sum_{(u,v) \in \gE}\left|\mX_u^{k} - \mX_v^{k} \right| \leq C_1 e^{-C_2 k}.
\end{equation*}

In real-world graphs, it is often the case that not all edges in a graph is positively curved. Nevertheless, we expect an abundance of edges with overly high curvature will either cause or worsen the over-smoothing issue in GNNs.

\subsection{Negative Graph Curvature and Over-squashing}

In this section, we demonstrate the intimate connection between negative graph curvature and the occurrence of local bottlenecks, which in turn causes over-squashing. 

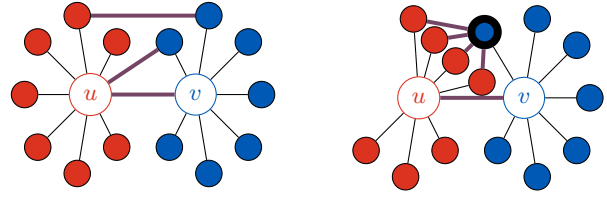
\begin{figure}[t]
\vskip 0.1in
\centering
\begin{subfigure}{0.45\columnwidth}
    \begin{tikzpicture}[scale = 0.7, u/.style = {draw, circle, red_colorblind, minimum size=1em},
    v/.style = {draw, circle, blue_colorblind, minimum size=1em},
    neighboru/.style = {draw, circle, fill = red_colorblind, minimum size=1em}, neighborv/.style = {draw, circle, fill = blue_colorblind, minimum size=1em},
    neighboruv/.style = {draw, circle, fill = violet_colorblind, minimum size=1em}] 
        \node[u , at={(0,0)}] (1) {\small$u$}; 
        \node[v , at={(2,0)}] (2) {\small $v$}; 
        \node[neighborv , at={(2.25,1.5)}] (3) {}; 
        \node[neighborv , at={(2.25,-1.5)}] (4) {}; 
        \node[neighboru , at={(-0.25,1.5)}] (5) {}; 
        \node[neighboru , at={(-0.25,-1.5)}] (6) {}; 
        \node[neighboru , at={(0.5,1)}] (7) {}; 
        \node[neighboru , at={(0.5,-1)}] (8) {}; 
        \node[neighboru , at={(-1,1)}] (9) {}; 
        \node[neighboru , at={(-1,-1)}] (10) {};
        \node[neighboru , at={(-1.25,0)}] (15) {}; 
        \node[neighborv , at={(1.5,1)}] (11) {}; 
        \node[neighborv , at={(1.5,-1)}] (12) {}; 
        \node[neighborv , at={(3,1)}] (13) {}; 
        \node[neighborv , at={(3,-1)}] (14) {}; 
        \node[neighborv , at={(3.25,0)}] (16) {}; 
        
        \draw (2) -- (3);
        \draw (2) -- (4);
        \draw (1) -- (5);
        \draw (1) -- (6);
        \draw (1) -- (7);
        \draw (1) -- (8);
        \draw (1) -- (9);
        \draw (1) -- (10);
        \draw (1) -- (15);
        \draw (2) -- (11);
        \draw (2) -- (12);
        \draw (2) -- (13);
        \draw (2) -- (14);
        \draw (2) -- (16);
        \draw[violet_colorblind, line width = 1.5pt] (1) -- (2);
        \draw[violet_colorblind, line width = 1.5pt] (3) -- (5);
        \draw[violet_colorblind, line width = 1.5pt] (11) -- (1);
    \end{tikzpicture} 
    \caption{Violet edges connecting $\Tilde{\mathcal{N}}_u \backslash \{v\}$ and $\Tilde{\mathcal{N}}_v \backslash \{u\}$ serve as information pathways.}
    \label{fig:negave_curvature_edges}
\end{subfigure}
\hfill
\begin{subfigure}{0.45\columnwidth}
     \begin{tikzpicture}[scale = 0.7, u/.style = {draw, circle, red_colorblind, minimum size=1em},
    v/.style = {draw, circle, blue_colorblind, minimum size=1em},
    neighboru/.style = {draw, circle, fill = red_colorblind, minimum size=1em}, neighborv/.style = {draw, circle, fill = blue_colorblind, minimum size=1em},
    neighboruv/.style = {draw, circle, fill = violet_colorblind, minimum size=1em}] 
        \node[u , at={(0,0)}] (1) {\small $u$}; 
        \node[v , at={(2,0)}] (2) {\small $v$}; 
        \node[neighborv , at={(2.25,1.5)}] (3) {}; 
        \node[neighborv , at={(2.25,-1.5)}] (4) {}; 
        \node[neighboru , at={(0.7,0.7)}] (5) {}; 
        \node[neighboru , at={(-0.25,-1.5)}] (6) {}; 
        \node[neighboru , at={(1.2,0.3)}] (7) {}; 
        \node[neighboru , at={(0.5,-1)}] (8) {}; 
        \node[neighboru , at={(0.3,1.1)}] (9) {}; 
        \node[neighboru , at={(-1,-1)}] (10) {};
        \node[neighboru , at={(-0.1,1.5)}] (15) {}; 
        \node[neighborv , at={(1.25,1.25)}, line width = 3] (11) {}; 
        \node[neighborv , at={(1.5,-1)}] (12) {}; 
        \node[neighborv , at={(3,1)}] (13) {}; 
        \node[neighborv , at={(3,-1)}] (14) {}; 
        \node[neighborv , at={(3.25,0)}] (16) {}; 
        
        \draw (2) -- (3);
        \draw (2) -- (4);
        \draw (1) -- (5);
        \draw (1) -- (6);
        \draw (1) -- (7);
        \draw (1) -- (8);
        \draw (1) -- (9);
        \draw (1) -- (10);
        \draw (1) -- (15);
        \draw (2) -- (11);
        \draw (2) -- (12);
        \draw (2) -- (13);
        \draw (2) -- (14);
        \draw (2) -- (16);
        \draw[violet_colorblind, line width = 1.5pt] (1) -- (2);
        \draw[violet_colorblind, line width = 1.5pt] (11) -- (5);
        \draw[violet_colorblind, line width = 1.5pt] (11) -- (9);
        \draw[violet_colorblind, line width = 1.5pt] (11) -- (7);
        \draw[violet_colorblind, line width = 1.5pt] (15) -- (11);
    \end{tikzpicture} 
    \caption{Edges may exacerbate the situation by creating new bottlenecks.}
    \label{fig:ideal_set}
\end{subfigure}     
\vskip -0.05in
\caption{Identifying connections that enable effective  message-passing at local neighborhoods.}
\label{fig:negative_curvature_illustrations}
\vskip -0.1in
\end{figure}

Message-passing across local neighborhoods is facilitated by connections of the form $(p,q)$ with $p \in \Tilde{\mathcal{N}}_u \backslash \{v\}$ and $q \in \Tilde{\mathcal{N}}_v \backslash \{u\}$. As visualized by Figure \ref{fig:negave_curvature_edges}, such edges (colored in violet) provide information pathways between $\Tilde{\mathcal{N}}_u$ and $\Tilde{\mathcal{N}}_v$. However, a large number of {these edges} concentrated on a relatively few vertices will create new bottlenecks, instead of providing good message channels. Figure \ref{fig:ideal_set} illustrates this point, as there are way too many edges connecting to the emphasized node but too little edges connecting between other neighbors. Since $n \geq m$, there is a natural squashing of information as messages are transmitted from $\mathcal{N}_u$ to $\mathcal{N}_v$ of ratio $\frac{n}{m}$. We identify the edges that provide good pathways as those that do not exacerbate this ratio and restrict our attention to these edges.

We characterize the effect of edge curvature on graph bottlenecks in the following proposition.

\begin{proposition}
\label{prop:negative_curvature_bottleneck}
Let $\Tilde{\gE}$ be union of the edge set $\gE$ with the set of all possible self-loops. Let $S$ be the subset of $\Tilde{\gE}$ containing edges of the form $(p,q)$ with $p \in \Tilde{\mathcal{N}}_u \backslash \{v\}$ and $q \in \Tilde{\mathcal{N}}_v \backslash \{u\}$. Supposing each vertex $w$ is a vertex of at most $\frac{n}{m}$ edges in $S$. The following inequality holds
\begin{equation}
    |S| \leq \frac{n (\kappa(u,v) + 2)}{2} .
\end{equation}
\end{proposition}

Recall that the curvature is deemed very negative if it is close to $-2$. Proposition \ref{prop:negative_curvature_bottleneck} shows that very negative edge curvature values prohibit the number of information pathways from $\Tilde{\mathcal{N}}_u$ to $\Tilde{\mathcal{N}}_v$, and very negatively curved edges induce local bottlenecks. This, in turn, contributes to the occurrence of the over-squashing issue as proposed by \cite{alon2021}. We note that an adequate measure for the over-squashing issue is currently lacking in the literature (see Appendix \ref{sec:over_squashing_measure}). Inspired by the influence distribution introduced by Xu et al. \yrcite{xu2018}, the next theorem asserts that negative edge curvature directly causes the decaying importance of distant nodes in GNNs with non-linearity removed. This demonstrates the effect of edge curvature on the over-squashing issue. \par

\begin{theorem}
\label{thm:negative_curvature_oversquashing}
Consider the update rule given by \Eqref{eq:general_mpnn}. Suppose $\psi_k$, $\phi_k$ are linear operators for all $k$, and $\bigoplus$ is the sum operation. If $u, v$ are neighboring vertices with neighborhoods as in Proposition \ref{prop:negative_curvature_bottleneck} and $S$ is defined similarly then for all $p \in \Tilde{\mathcal{N}}_u \backslash \{v\}$, $q \in \Tilde{\mathcal{N}}_v \backslash \{u\}$, we have
\begin{equation}
\label{eq:negative_curvature_oversquashing1}
    \begin{split}
        & \left[\frac{\partial \mX_u^{k+2}}{\partial \mX_q^k}\right] = \alpha \sum_{w \in V} \left[\frac{\partial \mX_u^{k+2}}{\partial \mX_w^k}\right], \\
        & \left[\frac{\partial \mX_v^{k+2}}{\partial \mX_p^k}\right] = \beta \sum_{w \in V} \left[\frac{\partial \mX_v^{k+2}}{\partial \mX_w^k}\right],
    \end{split}
\end{equation}
where $\left[\frac{\vy}{\vx} \right]$ is used to denote the Jacobian of $\vy$ with regard to $\vx$, and $\alpha, \beta$ satisfy
\begin{equation}
    \label{eq:negative_curvature_oversquashing2}
    \begin{split}
        & \alpha 
        \leq \frac{|S| + 2}{\sum_{w \in \Tilde{\mathcal{N}}_v} (\deg(w) + 1)},\\
        & \beta \leq \frac{|S| + 2}{\sum_{w \in \Tilde{\mathcal{N}}_u} (\deg(w) + 1)}.
    \end{split}
\end{equation}
\end{theorem}
To understand the meaning of Theorem~\ref{thm:negative_curvature_oversquashing}, let us fix $k = 0$ and assume $\gG$ is a regular graph with node degree $n$. Equations (\ref{eq:negative_curvature_oversquashing1}) and (\ref{eq:negative_curvature_oversquashing2}), along with Proposition \ref{prop:negative_curvature_bottleneck}, say that the contribution by the vertex $q$ to the vertex $u$ relative to the contribution of all other vertices, measured by the scaling term $\alpha$ in regard to the Jacobians, is bounded by $$\alpha \leq \frac{n(\kappa(u,v) + 2) + 4}{2(n+1)^2}.$$ Hence, {if $n$ is large and connections between the two neighborhoods are sparse, then $\kappa(u,v)$ is very negative. In such a case, we expect that the messages from each node of $\mathcal{N}_v$ make up only $\frac{2}{(n+1)^2}$ of the total sum of information.} They thus hardly have any effect on $u$, even when the distance between them is only $2$. An analogous result holds for the case when $\bigoplus$ is the mean operation without much modification. 

We have thus shown that negative curvature characterizes graph bottlenecks. As such, GNNs that operate on graphs with a large volume of negatively curved edges are expected to suffer from over-squashing.

\section{BORF: Batch Ollivier-Ricci Flow}\label{sec:borf}
Our theoretical results suggest the strikingly simple geometric connection between the over-smoothing and over-squashing issues: over-smoothing happens when there is a large proportion of edges with very positive curvature, while over-squashing occurs when there is a large proportion of edges with very negative curvature. As a natural extension, we propose the Batch Ollivier-Ricci Flow (BORF), a graph rewiring algorithm capable of simultaneously mitigating these issues by suppressing the over-smoothing and alleviateing the over-squashing inducing graph edges (see Algorithm~\ref{alg:borf}).


For each of $N$ batches, BORF first finds the $h$ edges $(u_1, v_1), \dots, (u_h, v_h)$ with minimal curvature and $k$ edges $(u^1, v^1), \dots, (u^k, v^k)$ with maximal curvature within the graph. {Note that we index the minimally curved and maximally curved edges by subscripts and superscripts, respectively.} Then, it tries to uniformly alleviate graph bottlenecks by adding connections to the set of $h$ minimally curved edges. To save on computation time, BORF does not recalculate the graph curvature within each batch. Instead, for each edge with minimal curvature $(u_j, v_j)$, it reuses the already calculated optimal transport plan $\pi_j$ between $\mu_{u_j}$ and $\mu_{v_j}$ to decide which edge should be added. Recall that the formula for the optimal transport cost is 
$$W_1(\mu_{u_j}, \mu_{v_j}) = \sum_{(p,q)} \pi_j(p,q) d(p,q).$$ Hence, to minimize the transport cost, it makes sense to rewire the two nodes that contribute the most to this sum. Specifically, we choose to add to $\gG$ the edge $(p^\ast,q^\ast)$ such that
$$(p^\ast,q^\ast) = \operatorname{argmax} \pi_j(p,q) d(p,q).$$

If there are multiple candidates, we arbitrarily choose one. Finally, BORF removes the $k$ maximally curved edges $(u^1, v^1), \dots, (u^k, v^k)$ whose presence might prime the over-smoothing behavior to occur. 

\begin{algorithm}[t!]
   \caption{Batch Ollivier-Ricci Flow (BORF)}
   \label{alg:borf}
\begin{algorithmic}
   \STATE {\bfseries Input:} graph $\gG = (\gV, \gE)$, \# rewiring batches $N$, \# edges added per batch $h$, \# edges removed per batch $k$
   \FOR{$i = 1$ {\bfseries to} $N$}
    \STATE Find $h$ edges $(u_1, v_1), \dots, (u_h,v_h)$ with minimal Ollivier-Ricci curvature $\kappa$, along with each summand $\pi_j(p,q) d(p,q)$ in their optimal transportation cost sum for all $p,q \in \gV$ and $j = \overline{1,h}$
    \STATE Find $k$ edges $(u^1, v^1), \dots, (u^k,v^k)$ with maximal Ollivier-Ricci curvature $\kappa$
    \FOR{$j = 1$ {\bfseries to} $h$}
    \STATE Add to $\gG$ the edge $(p^\ast,q^\ast)$ given by $$ (p^\ast,q^\ast) = \operatorname{argmax} \pi_j(p,q) d(p,q) $$
    \ENDFOR
    \STATE Remove edges $(u^1, v^1), \dots, (u^k,v^k)$ from $\gG$
   \ENDFOR
\end{algorithmic}
\end{algorithm}

With such design, BORF can effectively limit both ends of the curvature spectrum, simultaneously suppressing over-smoothing and {alleviating} over-squashing inducing connections. Furthermore, depending on data characteristics, we may change the behaviours of BORF to either be a net edge add, net edge minus, or net zero rewiring algorithm. Thereby, BORF permits fine-tuned and fluid adjustments of the total number of edges and their curvature range. 

{\textbf{Computational complexity.} Let $R$ be the number of edges to be rewired, $E = |\gE|$ be the number of edges in the graph, $D$ be the maximal degree of vertices on the graph, and $h+k$ be the number of edges rewired per batch by BORF. Solving the transportation problem to calculate the curvature for each edge using the network simplex algorithm is known to run in cubic time $O(D^3)$ \cite{ahuja1993network}. Once we have calculated the curvature of an edge, the optimal transport plan for that edge is already available for free. Hence, for each batch, the computational cost is $O(ED^3)$. The number of batches is $R/(h+k)$. Thus, the total cost for BORF is $O\left(RED^3/(h+k) \right)$ for each graph in consideration. 

Many efficient methods for approximating the Wasserstein distance, which is the computational bottleneck in BORF, have been proposed. In particular, log-linear and linear time approximations for the Wasserstein distance have recently been developed in \cite{bonneel2014sliced, atasu2019linear}, and some of them are already known to be topologically equivalent to the Wasserstein distance itself under appropriate assumptions \cite{erhan2021strong}. By incorporating these approximations to calculate the Ollivier-Ricci curvature, we may significantly reduce the computational cost of BORF. We leave studying such modifications as directions for future work.}

\textbf{Other rewiring algorithms.} SDRF \cite{topping2022} and FoSR \cite{karhadkar2022} are state-of-the-art rewiring algorithms for GNNs, designed with the purpose of alleviating the over-squashing issue. SDRF is based on the edge metric Balanced Forman curvature (BFC), which is actually a lower bound for the Ollivier-Ricci curvature. At its core, SDRF iteratively finds the edge with the lowest BFC, calculate the change in BFC for every possible edge that can be added, then add the one edge that affects the greatest change to the BFC of the aforementioned edge. On the other hand, FoSR is based on the heuristics that the spectral gap characterizes the connectivity of a graph. At each step, it approximates which missing edge would maximally improve the spectral gap and add that edge to the graph.

\textbf{Comparison to SDRF.} BORF shares some similarities to SDRF, but with notable differences. Since SDRF is based on BFC, it can only accurately enforce a lower bound on the Ollivier-Ricci curvature. Furthermore, its design lacks the capability to be used as a net edge minus rewiring algorithm.  As such, it is ill-equipped to deal with the over-smoothing issue or to be used on denser graphs. Another difference is BORF calculates the graph curvature very infrequently, while SDRF has to constantly recalculate for each possible new edge. Finally, by rewiring edges in batch, BORF affects a uniform change across the graph. This helps to preserve the graph topology {and prevent} the possibility that a small outlier subgraph gets continually rewired, while other parts of the graph do not see any geometrical improvement.

\textbf{Comparison to FoSR.} Unlike BORF and SDRF, FoSR does not have the ability to remove edges. Hence, despite over-smoothing and over-squashing being problems on the two ends of the same spectrum, the algorithm is incapable of addressing the first issue. It is also very challenging to predict where new edges will be added and what changes FoSR would make to the graph topology. This might complicate attempts by users to analyse the performance changes made by the algorithm.

\section{Experiments} \label{sec:experiments}

In this section, we empirically verify the effectiveness of BORF on a variety of tasks compared to other rewiring alternatives. We seek to demonstrate the potential of curvature-based rewiring methods, and more generally, geometric-aware techniques in improving the performance of GNNs. Our codes for the experiments are available at \url{https://github.com/hieubkvn123/revisiting-gnn-curvature}. 

\begin{table*}[ht!]
\caption{Classification accuracies of GCN and GIN with None, SDRF, FoSR, and BORF rewiring on various node classification datasets. Best results are highlighted in bold.}
\vskip -0.1in
\label{table:node_classification_results}
\begin{center}
\resizebox{\linewidth}{!}{
\begin{small}
\begin{sc}
\begin{tabular}{lcccc cccc}
\toprule
& \multicolumn{4}{c}{GCN}& \multicolumn{4}{c}{GIN}\\
\cmidrule(lr){2-5}
\cmidrule(lr){6-9}

Data set & None & SDRF & FoSR & BORF & None & SDRF & FoSR & BORF \\
\midrule
Cora &
  $86.7 \pm 0.3$ &
  $86.3 \pm 0.3$ &
  $85.9 \pm 0.3$ &
  \boldsymbol{$87.5 \pm 0.2$} &
  $76.0 \pm 0.6$ &
  $74.9 \pm 0.1$ &
  $75.1 \pm 0.8$ &
  \boldsymbol{$78.4 \pm 0.4$} \\
Citeseer & 
  $72.3 \pm 0.3$ &
  $72.6 \pm 0.3$ &
  $72.3 \pm 0.3$ &
  \boldsymbol{$73.8 \pm 0.2$} &
  $59.3 \pm 0.9$ &
  $60.3 \pm 0.8$ &
  $61.7 \pm 0.7$ &
  \boldsymbol{$63.1 \pm 0.8$}\\
Texas &
  $44.2 \pm 1.5$&
  $43.9 \pm 1.6$ &
  $46.0 \pm 1.6$&
  \boldsymbol{$49.4 \pm 1.2$}&
  $53.5 \pm 3.1$ &
  $50.3 \pm 3.7$ &
  $47.0 \pm 3.7$ &
  \boldsymbol{$63.1 \pm 1.7$}
  \\
Cornell &
  $41.5 \pm 1.8$ &
  $42.2 \pm 1.5$ &
  $40.2 \pm 1.6 $&
  \boldsymbol{$50.8 \pm 1.1$}&
  $36.5 \pm 2.2$ &
  $40.0 \pm 2.1$ &
  $35.6 \pm 2.4$ &
  \boldsymbol{$48.6 \pm 1.2$} \\
Wisconsin &
  $44.6 \pm 1.4$ &
  $46.2 \pm 1.2$ &
  $48.3 \pm 1.3$ &
  \boldsymbol{$50.3 \pm 0.9$}&
  $48.5 \pm 2.2$ &
  $48.8 \pm 1.9$ &
  $48.5 \pm 2.1$ &
  \boldsymbol{$54.9 \pm 1.2$} \\
Chameleon &
  $59.2 \pm 0.6$ &
  $59.4 \pm 0.5$ &
  $59.3 \pm 0.6$ &
  \boldsymbol{$61.5 \pm 0.4$}&
  $58.1 \pm 2.1$ &
  $58.4 \pm 2.1$ &
  $56.3 \pm 2.2$ &
  \boldsymbol{$65.3 \pm 0.8$}\\
\bottomrule
\end{tabular}
\end{sc}
\end{small}
}
\end{center}
\vskip -0.15in
\end{table*}

\begin{table*}[ht!]
\caption{Classification accuracies of GCN and GIN with None, SDRF, FoSR, and BORF rewiring on various graph classification datasets. Best results are highlighted in bold.}
\vskip -0.1in
\label{table:graph_classification_results}
\begin{center}
\resizebox{\linewidth}{!}{
\begin{small}
\begin{sc}
\begin{tabular}{lcccccccc}
\toprule
& \multicolumn{4}{c}{GCN} & \multicolumn{4}{c}{GIN} \\
\cmidrule(lr){2-5}
\cmidrule(lr){6-9}
Data set & None & SDRF & FoSR & BORF & None & SDRF & FoSR & BORF \\
\midrule
Enzymes &
  $25.5 \pm 1.3$ &
  $26.1 \pm 1.1$ &
  \boldsymbol{$27.4 \pm 1.1$} &
  $24.7 \pm 1.0$ &
  $31.3 \pm 1.2$ &
  $33.5 \pm 1.3$ &
  $25.3 \pm 1.2$ &
  \boldsymbol{$35.5 \pm 1.2$} \\
Imdb & 
  $49.3 \pm 1.0$ &
  $49.1 \pm 0.9$ &
  $49.6 \pm 0.8$ &
  \boldsymbol{$50.1 \pm 0.9$} &
  $69.0 \pm 1.3$ &
  $68.6 \pm 1.2$ &
  $69.5 \pm 1.1$ &
  \boldsymbol{$71.3 \pm 1.5$} \\
Mutag &
  $68.8 \pm 2.1$ &
  $70.5 \pm 2.1$ &
  $75.6 \pm 1.7$ &
  \boldsymbol{$75.8 \pm 1.9$} &
  $75.5 \pm 2.9$ &
  $77.3 \pm 2.3$ & 
  $75.2 \pm 3.0$ &
  \boldsymbol{$80.8 \pm 2.5$} \\
Proteins &
  $70.6 \pm 1.0$ & 
  $71.4 \pm 0.8$ &
\boldsymbol{$72.3 \pm 0.9$}&
  $71.0 \pm 0.8$ &
  $69.7 \pm 1.0$ &
  $72.2 \pm 0.9$ &
  \boldsymbol{$74.2 \pm 0.8$}& 
  $71.3 \pm 1.0$ \\
\bottomrule
\end{tabular}
\end{sc}
\end{small}
}
\end{center}
\vskip -0.15in
\end{table*}

{\color{blue}
\begin{table*}[ht!]
\caption{Classification accuracies of GCN at depths 5, 7, and 9 with different BORF rewiring options on Cornell and Mutag datasets.}
\label{table:abridged_ablation_study_edge_addition_removal}
\begin{center}
\begin{small}
\begin{sc}
\begin{tabular}{lcccccc}
\toprule
Data set & \# layers & None & Best settings & Only remove & Only add & remove \& add equally\\
\midrule
 \multirow{3}{*}{Cornell} & 5 &
  $41.3 \pm 1.4$ &
  $45.5 \pm 1.1$ &
  $46.4 \pm 1.2$ &
  $44.7 \pm 1.3$ &
  $45.9 \pm 1.2$ \\
& 7 &
  $39.5 \pm 1.7$ &
  $41.5 \pm 1.5$ &
  $43.2 \pm 1.3$ &
  $42.8 \pm 1.4$ &
  $41.8 \pm 1.3$ \\
& 9 &
  $35.5 \pm 1.4$ &
  $40.9 \pm 1.3$ &
  $41.9 \pm 1.6$ &
  $40.3 \pm 2.0$ &
  $39.9 \pm 1.6$ \\
\midrule
\multirow{3}{*}{Mutag} & 5 &
  $67.7 \pm 1.6$ &
  $75.4 \pm 2.1$ &
  $68.5 \pm 2.8$ & 
  $76.1 \pm 2.2$ &
  $71.8 \pm 1.2$ \\
& 7 &
  $64.1 \pm 2.1$ &
  $72.1 \pm 1.3$ &
  $65.1 \pm 1.5$ & 
  $75.2 \pm 2.4$ &
  $66.2 \pm 1.9$ \\
& 9 &
  $63.1 \pm 1.2$ &
  $69.7 \pm 1.5$ &
  $60.7 \pm 2.5$ & 
  $70.4 \pm 1.7$ &
  $61.3 \pm 1.5$ \\
\bottomrule
\end{tabular}
\end{sc}
\end{small}
\end{center}
\vskip -0.15in
\end{table*}
}

\textbf{Datasets.} We conduct our experiments on a range of widely used node classification and graph classification tasks. For node classification, we report our results on the datasets CORA, CITESEER \cite{planetoid}, TEXAS, CORNELL, WISCONSIN \cite{webk_dataset} and CHAMELEON \cite{wikipedia_dataset}. For graph classification, we {validate our method} on the following benchmarks: ENZYMES, IMDB, MUTAG and PROTEINS from the TUDataset \cite{tudataset}. A summary of dataset statistics is available in Appendix \ref{sec:dataset_statistics}.

\textbf{Experiment details.} {We compare BORF to no graph rewiring and two other state-of-the-art rewiring methods: SDRF \cite{topping2022} and FoSR \cite{karhadkar2022}}. 
{In designing our experiments, we prioritize fairness and comprehensiveness, rather than aiming to obtain the best possible performance for each dataset presented.}  
We applied each method as a preprocessing step to all graphs in the datasets considered, before feeding the rewired graph data into a GNN to evaluate performance. For baseline GNNs, we employed the popular graph architectures GCN \cite{kipf2016} and GIN \cite{xu2019}. 
For each task and baseline model, we used the same {settings} of GNN and optimization hyper-parameters across all rewiring methods to rule out hyper-parameter tuning as a source of performance gain. The setting for each rewiring option was obtained by tuning every hyper-parameter available for each method with the exception of the temperature $\tau$ of SDRF, which we set to $\infty$. Each configuration is evaluated using the validation set. The test set accuracy of the configuration with the best validation performance is then recorded. For each experiment, we accumulate the result across $100$ random trials and report the mean test accuracy, along with the $95\%$ confidence interval. Further experiment details are available in Appendix \ref{sec:experiment_settings}.

\textbf{Results. } 
Table \ref{table:node_classification_results} and Table \ref{table:graph_classification_results} summarize our experiment results for node classification and graph classification datasets, respectively. BORF outperforms all other methods in every node classification tasks on both GCN and GIN. It is worth mentioning that the $95\%$ confidence interval of BORF is almost always smaller than other methods, indicating a consistent level of performance. This result {agrees} with what is expected since SDRF and FoSR are not suited for dealing with the over-smoothing issue, which heavily {degrades the model's performance} on node classification tasks. {On graph classification datasets, BORF achieves higher test accuracy compared to other rewiring options in most settings.}

{\textbf{Ablation study on edge addition/removal.} To investigate the role of edge addition and removal, we compare BORF's performance on GCN at high depths when using the best rewiring settings found in previous experiments against versions of BORF where edges are only removed (only alleviates over-smoothing), edges are only added (only alleviates over-squashing), or edges are removed and added equally. Further experimental details are described in Section \ref{sec:edge_addition_removal}. We report our experimental results in Table \ref{table:abridged_ablation_study_edge_addition_removal}.

At all depths, the GNN can learn relevant information from the datasets preprocessed with BORF more effectively, even when the rewiring algorithm is constrained to only alleviate either over-smoothing or over-squashing. Indeed, the un-rewired datasets record the worst performance in 5 out of 6 scenarios considered. This indicates that both issues negatively impact the classification accuracy as the depth count increases. Our results suggest both BORF's abilities to reduce over-smoothing and over-squashing play an essential role in improving the performance of GNN models.
}

\section{Conclusion}\label{sec:conclusion}
In this paper, we established the novel correspondence between the Ollivier-Ricci curvature on graphs with both the over-smoothing and the over-squashing issues. In specific, we showed that positive graph curvature is associated with over-smoothing, while negative graph curvature is associated with over-squashing. Based on our theoretical results, we proposed Batch Ollivier-Ricci Flow, a novel curvature based rewiring method that can effectively improve GNN performance by tackling both the over-smoothing and over-squashing problems at the same time. It is interesting to note that by different definitions of the random walk $\mu_u$, we may be able to capture different behaviors of the local graph structures using graph curvature. 

{
\textbf{Limitations and societal impacts.} Our current formulation only demonstrates that the over-squashing issue is related to negative Ollivier-Ricci curvature at a local scale under some constraints on the graph structure and GNN design. We leave providing a more general treatment of the problem for future work. Our theory and proposed method have no discernible negative societal impact.}


\section*{Acknowledgements}
This material is based on research sponsored by the NSF Grant\# 2030859 to the Computing Research Association for the CIFellows Project (CIF2020-UCLA-38). SJO acknowledges support from the 
ONR N00014-20-1-2093/N00014-20-1-2787 and the NSF DMS 2208272 and 1952339.



\bibliography{references}
\bibliographystyle{icml2023}

\newpage
\appendix
\onecolumn
\begin{center}
{\bf \Large Supplement for ``Revisiting Over-smoothing and Over-squashing using Ollivier-Ricci Curvature"}
\end{center}
In this supplementary material, we first present how \Eqref{eq:general_mpnn} accommodates different designs of GNNs in Appendix \ref{sec:mpnn}. Then, we discuss the lack of an appropriate measure for the over-squashing issue in Appendix \ref{sec:over_squashing_measure}. Skipped proofs within the main text are provided in Appendix \ref{sec:Proofs}. We present our experiment settings in Appendix \ref{sec:experiment_settings} and additional experiment results in Appendix \ref{sec:additional_experiments}. Dataset statistics are available in Appendix \ref{sec:dataset_statistics}. Finally, we provide our hardware specifications and list of libraries in Appendix \ref{sec:hardware_library}.

\section{Message Passing Neural Networks}
\label{sec:mpnn}
In its most general form, a typical layer of a message passing neural network is given by the following update rule~\cite{bronstein2021}:
\begin{equation*} \label{eq:mpflavor}
        \mH_u = \phi \left(\mX_u, \bigoplus_{v \in \mathcal{N}_u} \Lambda(\mX_u, \mX_v) \right).
\end{equation*}
Here, $\Lambda$ , $\bigoplus$ and $\phi$ are the message, aggregate and update functions. Different designs of MPNNs amount to different choices for these functions. The additional input of $\mX_u$ to $\phi$ represents an optional skip-connection. 

In practice, we found that the skip connection of each vertex $u$ is often implemented by considering a message passing scheme where each node sends a message to itself. This can be thought of as adding self-loops to the graph, and its impact has been studied by Xu et al. \yrcite{xu2018} and Topping et al., \yrcite{topping2022}. Then, $\Lambda$ could be realized as a learnable affine transformation $\psi$ of $\mX_v$, the aggregating function $\bigoplus$ could either be chosen as a sum, mean, weighted mean, or max operation, and $\phi$ is a suitable activation function. Hence, we arrive at \Eqref{eq:general_mpnn}, which we restate below
\begin{equation*}
    \mX^{k+1}_u = \phi_k \left(\bigoplus_{p \in \Tilde{\mathcal{N}}_u} \psi_k(\mX^k_p) \right).
\end{equation*}
\par For example, graph convolutional network (GCN) \cite{kipf2016} defines its layer as
\begin{equation*}
    \mH_u = \sigma \left( \sum_{v \in \Tilde{\mathcal{N}}_u} \frac{1}{c_{uv}} \mW \mX_v \right),
\end{equation*}
with  $c_{uv} = \sqrt{|\Tilde{\mathcal{N}}_u||\Tilde{\mathcal{N}}_v|}$. The mean aggregator variant (but not other variants) of GraphSAGE \cite{hamilton2017} uses the same formulation but with $c_{uv} = |\Tilde{\mathcal{N}}_u|$. Both choices of $c_{uv}$ have the exact same spectral characteristics and act identically in theory \cite{li2018}, which leads to an averaging behavior based on node degrees. Similarly, graph attention networks (GAT) \cite{velickovic2018} defines its single head layer as
\begin{equation*}
    \mH_u = \sigma \left( \sum_{v \in \Tilde{\mathcal{N}}_u} a_{uv} \mW \mX_v \right).
\end{equation*}
The difference here being $a_{uv}$ is now a learnable function given by the attention mechanism \cite{vaswani2017}. Finally, graph isomorphism network (GIN) \cite{xu2019} is formulated by 
\begin{equation*}
    \mH_u = \operatorname{MLP} \left( (1+\epsilon) \mX_u + \sum_{v \in \mathcal{N}_u} \mX_v \right),
\end{equation*}
where $\operatorname{MLP}$ is a multilayer perceptron. GIN achieves its best performance with the model GIN-$0$, in which $\epsilon$ is set to be $0$.

We remark that most nonlinear activation functions such as ReLU, Leaky ReLU, Tanh, Sigmoid, softmax, etc., has a simple and explicit Lipschitz-constant (which equals to $1$ more often than not) \cite{scaman2018}. 


\section{Measuring Over-squashing}
\label{sec:over_squashing_measure}
It is intuitive to think that if $N$ vertices in $\Tilde{\mathcal{N}}_u$ contribute to the feature representation of some vertex $u$ by the permutation invariant update rule \ref{eq:general_mpnn}, we should expect each such vertex to provide $\frac{1}{N}$ of the total contribution. If this is repeated over and over, such as in a tree, the exponentially decaying dependence of distant vertices is to be expected. However, quantifying this phenomenon is actually quite difficult. The first work to call attention to the over-squashing problem \cite{alon2021} measures whether breaking the bottleneck improves the results of long-range problems, instead of measuring the decaying importance itself.

A definition for the over-squashing issue that took inspiration from the vanishing gradient problem was given by \cite{topping2022}, where it was suggested that $\frac{\partial \mX_u^k}{\partial \mX_v^0}$ can be used to evaluate the decreasing importance of distant vertex $v$ to $u$. {However, this quantity does not actually measure the squashing behavior experienced by all GNNs, but only by those using an aggregation function with a natural decaying effect such as GCNs. Furthermore, we argue that in Theorem 4 of \cite{topping2022}, there is no general squashing behavior being described. Given some negatively curved edge $(i,j)$, the conclusion of the theorem provides a bound for the average effect of messages from the node $i$ to a set of neighbors $Q_j$ of the node $j$. In other words, this describes how one single node on one specific side of the edge (the side with the lower degree) can transmit messages to other nodes and not the other way around. To visualize this, let us look at Figure \ref{fig:negave_curvature_edges}. The theorem measures how effectively the red node $u$ can send messages to a number of blue nodes in the neighborhood of $v$. Clearly, there is no squashing of information from the surrounding receptive field being considered: the message was sent from one single node, and there is no other message to be squashed. In reality, this theorem is describing the average dilution of information sent out by the node $i$ to other nodes $k$ of distance $2$ from $i$. This dilution is caused by the choice of using the normalized adjacency matrix as the aggregating function. If we use the sum aggregating function as in GIN, then such diluting process does not take place.} 
 
 Clearly, it is the \textit{relative} importance of a vertex compared to the contribution of all other vertices that is at the heart of the matter. To this end, we have found that the closest notion to our description actually predates the observation of the over-squashing issue. \citet{xu2018} introduced the notion of influence distribution to quantify the relative importance of vertices to each other. It is defined as
\begin{equation*}
    I_u(v) = \frac{\operatorname{sum}\left( \left[ \frac{\partial \mX_u^k}{\partial \mX_v^0} \right] \right)}{\sum_{p \in V} \operatorname{sum}\left( \left[\frac{\partial \mX_u^k}{\partial \mX_p^0} \right]\right)}
\end{equation*}
where the sum is taken over all entries of each Jacobian matrix. Unfortunately, this definition is quite unwieldy to use in any sort of analysis. We would like to remark that the theoretical proofs in \cite{xu2018} are only partially correct. They have made a mistake by claiming 
\begin{equation*}
    \mathbb{E} \left(\frac{X_1}{\sum_{i=1}^n X_i}\right) = \frac{\mathbb{E}(X_1)}{\sum_{i=1}^n \mathbb{E}(X_i)}
\end{equation*}
for random variables $X_i$.


\section{Proofs} \label{sec:Proofs}
In this Appendix, we provide proofs for key results in the paper. We state without proof the following lemma, which is Lemma 4.1 in \cite{bourne2017}. 
\begin{lemma} \label{lemma:bourne2017}
Let $\mu_1, \mu_2$ be probability measures on a space $V$. Then there exists an optimal transport plan $\pi$ transporting $\mu_1$ to $\mu_2$ with the following property: For all $x \in V$ with $\mu_1(x) \leq \mu_2(x)$, we have $\pi(x,x) = \mu_1(x)$.
\end{lemma}
\subsection{Proof of Lemma~\ref{thm:positive_curvature}}
\begin{proof}
Without loss of generality, assume $n \geq m$. Let $\pi$ be an optimal transport plan between $\mu_u$ and $\mu_v$ satisfying the condition in Lemma \ref{lemma:bourne2017}. That is, $\pi(p,p) = \frac{1}{n}$ for all $p \in \mathcal{N}(u) \cap \mathcal{N}(v)$. We have
\begin{align*}
    W_1(\mu_u, \mu_v) 
    &= \sum_{p \in \mathcal{N}_u} \sum_{q \in \mathcal{N}_v} \pi(p,q) d(p,q) \\
    &= \sum_{\substack{ (p,q) \in \mathcal{N}_u \times \mathcal{N}_v \\ p \neq q}} \pi(p,q) d(p,q) + \sum_{p \in \mathcal{N}(u) \cap \mathcal{N}(v)} \pi(p,p) d(p,p).
\end{align*}
It is obvious that $d(p,p) = 0$ for any vertex $p$ and $d(p,q) \geq 1$ for any vertices $p \neq q$. We have 
\begin{align*}
    W_1(\mu_u, \mu_v) 
    &\geq \sum_{\substack{ (p,q) \in \mathcal{N}_u \times \mathcal{N}_v \\ p \neq q}} \pi(p,q) + 0 \\
    &= 1 - \sum_{p \in \mathcal{N}_u \cap \mathcal{N}_v} \pi(p,p)\\
    &= 1 - \frac{|\mathcal{N}_u \cap \mathcal{N}_v|}{n}.
\end{align*}
Hence, we have 
\begin{equation*}
    \frac{|\mathcal{N}_u \cap \mathcal{N}_v|}{\max(m,n)} \geq 1 - W_1(\mu_u, \mu_v) = \kappa(u,v).
\end{equation*}
\end{proof}

\subsection{Proof of Theorem~\ref{thm:onelayerGNN}}
As $\phi_k$ is $L$-Lipschitz, we have 
\begin{align}
    \left|\mX^{k+1}_u - \mX^{k+1}_v \right| 
    &= \left|\phi_k \left(\bigoplus_{p \in \Tilde{\mathcal{N}}_u} \psi_k(\mX^k_p) \right) - \phi_k \left(\bigoplus_{q \in \Tilde{\mathcal{N}}_v} \psi_k(\mX^k_q) \right) \right| \nonumber\\
    &\leq L \left|\bigoplus_{p \in \Tilde{\mathcal{N}}_u} \psi_k(\mX^k_p) - \bigoplus_{q \in \Tilde{\mathcal{N}}_v} \psi_k(\mX^k_q) \right|. \label{eq:oplus_operation}
\end{align}
Theorem \ref{thm:positive_curvature} tells us that 
\begin{equation*}
    |\Tilde{\mathcal{N}}_v \backslash \Tilde{\mathcal{N}}_u| \leq |\Tilde{\mathcal{N}}_u \backslash \Tilde{\mathcal{N}}_v| = n+1 - |\mathcal{N}_u \cap \mathcal{N}_v| -2 \leq n - n \kappa(u,v).
\end{equation*}
Hence, there are at most $\lfloor (1-\kappa(u,v)) n \rfloor$ vertices in the extended neighborhood of $u$ that is not present in the extended neighborhood of $v$ and vice versa. The symmetric difference $\Tilde{\mathcal{N}}_u \vartriangle \Tilde{\mathcal{N}}_v$ satisfies
$$|\Tilde{\mathcal{N}}_u \vartriangle \Tilde{\mathcal{N}}_v| = |(\Tilde{\mathcal{N}}_u \backslash \Tilde{\mathcal{N}}_v) \cup (\Tilde{\mathcal{N}}_v \backslash \Tilde{\mathcal{N}}_u)| \leq 2 (1-\kappa(u,v)) n.$$  
\begin{itemize}
    \item If $\bigoplus$ is realized as the sum operation, we obtain from Equation (\ref{eq:oplus_operation})
    \begin{align*}
        \left|\mX^{k+1}_u - \mX^{k+1}_v \right| 
        &\leq  L \left|\sum_{p \in \mathcal{N}_u \cup \{u\}} \psi_k(\mX^k_p) - \sum_{q \in \mathcal{N}_v \cup \{v\}} \psi_k(\mX^k_q) \right|\\
        &= L \left|\sum_{p \in \Tilde{\mathcal{N}}_u \backslash \Tilde{\mathcal{N}}_v} \psi_k(\mX^k_p) - \sum_{q \in \Tilde{\mathcal{N}}_v \backslash \Tilde{\mathcal{N}}_u} \psi_k(\mX^k_q) \right|\\
        &\leq L \sum_{p \in \Tilde{\mathcal{N}}_u \vartriangle \Tilde{\mathcal{N}}_v} \left| \psi_k(\mX_p^k) \right|\\
        &\leq (1-\kappa(u,v)) 2 L C M  n.
    \end{align*}
    We can now set $h \equiv 2LCMn$.
    \item If $\bigoplus$ is realized as the mean operation, we obtain from Equation (\ref{eq:oplus_operation})
    \begin{align}
        \left|\mX^{k+1}_u - \mX^{k+1}_v \right| 
        &\leq  L \left|\sum_{p \in  \Tilde{\mathcal{N}}_u} \frac{1}{n+1} \psi_k(\mX^k_p) - \sum_{q \in  \Tilde{\mathcal{N}}_v} \frac{1}{m+1} \psi_k(\mX^k_q) \right| \nonumber\\
        &\leq L \sum_{p \in ( \Tilde{\mathcal{N}}_u \cap  \Tilde{\mathcal{N}}_v)} \left(\frac{1}{m+1} - \frac{1}{n+1}\right) \left|\psi_k(\mX^k_p) \right| \nonumber\\
        & \hspace{0.5cm} +L \left|\sum_{p \in  \Tilde{\mathcal{N}}_u \backslash  \Tilde{\mathcal{N}}_v} \frac{1}{n+1} \psi_k(\mX^k_p) - \sum_{q \in  \Tilde{\mathcal{N}}_v \backslash  \Tilde{\mathcal{N}}_u} \frac{1}{m+1} \psi_k(\mX^k_q) \right| \label{eq:positive_curvature_1}.
    \end{align}
    We have $n \geq m  = |\mathcal{N}_v| \geq |\mathcal{N}_u \cap \mathcal{N}_v| \geq \kappa(u,v) n$, and
    $$\frac{1}{m+1} - \frac{1}{n+1} \leq \frac{1}{m} - \frac{1}{n} \leq \frac{1}{\kappa(u,v) n} - \frac{1}{n} = \frac{1 - \kappa(u,v)}{\kappa(u,v) n}.$$
    Therefore, Equation (\ref{eq:positive_curvature_1}) gives 
    \begin{align*}
    \left|\mX^{k+1}_u - \mX^{k+1}_v \right| 
        &\leq L \sum_{p \in \Tilde{\mathcal{N}}_u \cap  \Tilde{\mathcal{N}}_v} \frac{1-\kappa(u,v)}{\kappa(u,v) n} \left|\psi_k(\mX^k_p) \right| +L \sum_{p \in \Tilde{\mathcal{N}}_u \vartriangle \Tilde{\mathcal{N}}_v} \frac{1}{\kappa(u,v) n + 1} \left|\psi_k(\mX^k_p) \right| \\
        &\leq L (n+1) \frac{1-\kappa(u,v)}{\kappa(u,v) n} CM + 2 (1-\kappa(u,v)) n L \frac{1}{\kappa(u,v) n + 1} CM \\
        &\leq (1 - \kappa(u,v)) LCM  \left(\frac{n+1}{\kappa(u,v) n} + 2 \frac{n}{\kappa(u,v) n + 1}\right).
    \end{align*}
    We can now set $h(x) = LCM ( \frac{n+1}{x n} + 2 \frac{n}{nx+1})$.
\end{itemize}
Clearly, the functions $h$ as defined satisfy the conditions given in Theorem \ref{thm:onelayerGNN}.

\subsection{Proof of Proposition~\ref{prop:pos_curved_regular_multilayer_GNN}}
We will use proof by induction. For all edges $u \sim v$, repeating the argument in Theorem \ref{thm:onelayerGNN}, we get $|\Tilde{\mathcal{N}}_u \vartriangle \Tilde{\mathcal{N}}_v| \leq 2 (1-\delta) n$. Then, the base case $k = 1$ follows since
\begin{align*}
    |\mX^1_u - \mX^1_v| 
    &= \left|\phi_1\left( \frac{1}{n + 1} \sum_{p \in \Tilde{\mathcal{N}_u}}  \psi(\mX_p) \right) - \phi_1 \left( \frac{1}{n + 1} \sum_{q \in \Tilde{\mathcal{N}_v}} \psi(\mX_q) \right) \right|\\
    &\leq L \left|\frac{1}{n + 1} \sum_{p \in \Tilde{\mathcal{N}_u}}  \psi(\mX_p) - \frac{1}{n + 1} \sum_{q \in \Tilde{\mathcal{N}_v}} \psi(\mX_q) \right|
    \\
    &= \frac{L}{n + 1} \left| \sum_{p \in \Tilde{\mathcal{N}_u} \backslash \Tilde{\mathcal{N}_v}} \psi(\mX_p) - \sum_{q \in \Tilde{\mathcal{N}_v} \backslash \Tilde{\mathcal{N}_u}} \psi(\mX_q) \right|\\
    &\leq \frac{L}{n + 1}  \sum_{p \in \Tilde{\mathcal{N}}_u \vartriangle \Tilde{\mathcal{N}}_v} |\psi(\mX_p)| \\
    &\leq \frac{2 \lfloor (1-\delta) n \rfloor}{n+1} LCM.
\end{align*}

Suppose the statement is true for $k$ and consider the case $k+1$. We have for all $u \sim v$:
\begin{align}
    \left|\mX^{k+1}_u - \mX^{k+1}_v\right| \nonumber
    &\leq L\frac{1}{n + 1} \left| \sum_{p \in \Tilde{\mathcal{N}_u}} \psi_k(\mX^k_p) - \sum_{q \in \Tilde{\mathcal{N}_v}} \psi_k(\mX^k_q) \right|\\ 
    &= L\frac{1}{n + 1} \left| \sum_{p \in \Tilde{\mathcal{N}_u} \backslash \Tilde{\mathcal{N}_v}} \psi_k(\mX^k_p) - \sum_{q \in \Tilde{\mathcal{N}_v} \backslash \Tilde{\mathcal{N}_u}} \psi_k(\mX^k_q) \right| \nonumber \\
    &= L \frac{1}{n + 1}  \left| \psi_k \left( \sum_{p \in \Tilde{\mathcal{N}_u} \backslash \Tilde{\mathcal{N}_v}} \mX^k_p - \sum_{q \in \Tilde{\mathcal{N}_v} \backslash \Tilde{\mathcal{N}_u}} \mX^k_q \right)\right| \nonumber \\
    & \leq LM \frac{1}{n + 1}  \left| \sum_{p \in \Tilde{\mathcal{N}_u} \backslash \Tilde{\mathcal{N}_v}} \mX^k_p - \sum_{q \in \Tilde{\mathcal{N}_v} \backslash \Tilde{\mathcal{N}_u}} \mX^k_q \right|.\label{eq:multilayerGCN1}
\end{align}
For each $p \in \Tilde{\mathcal{N}_u} \backslash \Tilde{\mathcal{N}_v}$, match it with one and only one $q \in \Tilde{\mathcal{N}_v} \backslash \Tilde{\mathcal{N}_u}$. For any node pair $(p,q)$, they are connected by the path $p \sim u \sim v \sim q$, where the difference in norm of features at layer $k$ of each $1$-hop connection is at most $\frac{2}{3} C \left(\frac{3LM\lfloor(1-\delta) n \rfloor}{n+1}\right)^{k}$. Hence, we have 
$$|\mX^{k}_p - \mX^{k}_q| \leq 2 C \left(\frac{3LM\lfloor(1-\delta) n \rfloor}{n+1}\right)^{k}.$$
Substituting this into equation (\ref{eq:multilayerGCN1}), and by noting that there are at most $\lfloor (1-\delta) n \rfloor$ pairs, we get
\begin{align*}
    |\mX^{k+1}_u - \mX^{k+1}_v| 
    &\leq LM \frac{1}{n + 1} \sum_{(p,q)} \left|\mX_p^k - \mX_q^k \right|\\ 
    &\leq LM \frac{1}{n + 1} \lfloor(1-\delta) n \rfloor 2 C \left(\frac{3LM\lfloor(1-\delta) n \rfloor}{n+1}\right)^{k}\\
    &= \frac{2}{3} C \left(\frac{3LM\lfloor(1-\delta) n \rfloor}{n+1}\right)^{k+1}.
\end{align*}
By induction, we have shown inequality (\ref{eq:multilayerGNNbound}) holds for all $k \geq 1$ and $u \sim v$. 

It is known that if the curvature of every edge in a graph is positive and bounded away from zero by $\delta > 0$ then the diameter of the graph does not exceed $\lfloor2/\delta \rfloor$ \cite{Paeng2012VolumeAD}. Hence, for any two nodes $u,v \in \gV$, the shortest path between them is of length at most $\lfloor2/\delta \rfloor$. Apply the inequality (\ref{eq:multilayerGNNbound}) for each pair of neighboring nodes on this shortest path, we obtain the inequality (\ref{eq:multilayerGNNbound2}).

\subsection{Proof of Proposition~\ref{prop:negative_curvature_bottleneck}}
Note that $S$ consists of elements of the form $(p,q)$ where either $p = q$ or $p \neq q$. The first type corresponds to mutual neighbors of $u, v$, while the second type corresponds to neighbors of $u, v$ that share an edge. Denote the number of edges of the first type as $n_0$ and the number of edges of the second type as $n_1$. A transport plan $\pi$ between $\mu_u$ and $\mu_v$ can be obtained as followed.
\begin{itemize}
    \item For every vertex $p$ such that $(p,p) \in S$, the mass of $\mu_u(p) = \frac{1}{n}$ remains in place at $p$ with cost $\pi(p,p) \times 0 = \frac{1}{n} \times 0 = 0$
    \item For each edge $(p,q) \in S$ with $p \neq q$, transport the mass of $\frac{1}{n}$ from $p$ to $q$ with cost $\pi(p,q) \times 1 = \frac{1}{n} \times 1 = \frac{1}{n}$. The assumption that each vertex $w$ is a vertex of at most $\frac{n}{m}$ edges ensures that the total mass transported to each vertex is no greater than $\frac{1}{m}$.
    \item The remaining mass is $1 - n_0 \frac{1}{n} - n_1 \frac{1}{n}$. Transport this amount arbitrarily to obtain a valid optimal transport plan.
\end{itemize}
This transport plan has cost 
\begin{equation*}
   \sum_{p \in \Tilde{\mathcal{N}_u}} \sum_{q \in \Tilde{\mathcal{N}_v}} \pi(p,q) d(p,q) \leq  n_0 0 + n_1 \frac{1}{n} + \left(1 - n_0 \frac{1}{n} - n_1 \frac{1}{n}\right) 3 = 3 - 3n_0 \frac{1}{n} -2n_1 \frac{1}{n}. 
\end{equation*}
We have
\begin{equation*}
    \kappa(u,v) = 1 - W_1(\mu_u, \mu_v) \geq 1 - \left( 3 - 3n_0 \frac{1}{n} -2n_1 \frac{1}{n} \right) = -2 + \frac{3n_0 + 2n_1}{n}.
\end{equation*}
Therefore, we obtain
\begin{equation*}
    |S| = n_0 + n_1 \leq \frac{n}{2}  \frac{3n_0 + 2n_1}{n} \leq n \frac{\kappa(u,v) + 2}{2}.
\end{equation*}
We can observe from the proof that a stronger result holds: $3 n_0 + 2 n_1 \leq n (\kappa(u,v)+2)$.

\subsection{Proof of Theorem~\ref{thm:negative_curvature_oversquashing}}
Since $\phi_k$ and $\psi_k$ are linear operators for all $k$, their Jacobians $J_{\phi_{k}}, J_{\psi_{k}}$ are constant matrices. By inspection of \Eqref{eq:general_mpnn}, we see that a vertex $w \in V$ can only transmit a message to $u$ if there exists a vertex $w'$ such that $w' \in \Tilde{\mathcal{N}}_u \cap \Tilde{\mathcal{N}}_w$. Moreover, the chain rule gives
\begin{equation*}
    \left[ \frac{\partial \mX_u^{k+2}}{\partial \mX_w^k} \right] = \sum_{w' \in \Tilde{\mathcal{N}}_u \cap \Tilde{\mathcal{N}}_w} \left[ \frac{\partial \mX_u^{k+2}}{\partial \mX_{w'}^{k+1}} \right] \left[ \frac{\partial \mX_{w'}^{k+1}}{\partial \mX_{w}^{k}} \right] = \sum_{w' \in \Tilde{\mathcal{N}}_u \cap \Tilde{\mathcal{N}}_w} J_{\phi_{k+1}} J_{\psi_{k+1}} J_{\phi_{k}} J_{\psi_{k}}.
\end{equation*}
Therefore, $\left[ \frac{\partial \mX_u^{k+2}}{\partial \mX_w^k} \right]$ is the number of distinct paths (that might contain self-loops) from $w$ to $u$ times $J_{\phi_{k+1}} J_{\psi_{k+1}} J_{\phi_{k}} J_{\psi_{k}}$. \par 
The number of distinct paths without self-loops from $q \in \Tilde{\mathcal{N}}_v \backslash \{u\}$ to $u$ is not greater than $|S|$ as defined in Proposition \ref{prop:negative_curvature_bottleneck}. With self-loops, this rises to at most $|S|+2$, which corresponds to the case where $q \sim u$. \par 
On the other hand, $\sum_{w \in V} \left[\frac{\partial \mX_u^{k+2}}{\partial \mX_w^k}\right]$ equals the number of distinct paths with self-loops with one end at $u$ times $J_{\phi_{k+1}} J_{\psi_{k+1}} J_{\phi_{k}} J_{\psi_{k}}$. Clearly, we have 
\begin{equation*}
    \sum_{w \in V} \left[\frac{\partial \mX_u^{k+2}}{\partial \mX_w^k}\right] = \left(\sum_{w \in \Tilde{\mathcal{N}}_u} (\deg(w) + 1)\right) J_{\phi_{k+1}} J_{\psi_{k+1}} J_{\phi_{k}} J_{\psi_{k}}.
\end{equation*}
Let $$\alpha = \frac{|S|+2}{\sum_{w \in \Tilde{\mathcal{N}}_u} (\deg(w) + 1)},$$ then Proposition \ref{prop:negative_curvature_bottleneck} gives us the required inequality. We can choose $\beta$ by the same process.

\section{Experiment Settings}
\label{sec:experiment_settings}


\subsection{Rewiring hyper-parameters}
\label{sec:hyperparams}

\begin{table}[t!]
\centering
\caption{SDRF's best hyper-parameter settings.} \label{table:sdrf_hyperparams}
\vskip 0.15in
\begin{sc}
    \begin{small}
        \begin{tabular}{lcccccc}
\toprule
& \multicolumn{3}{c}{GCN} & \multicolumn{3}{c}{GIN}\\
\cmidrule(lr){2-4}
\cmidrule(lr){5-7}
Dataset &  \# iteration &  $C^+$ &  \# rewired & \# iteration &  $C^+$ &  \# rewired \\
\midrule
CORA      & $12$    & $0$  & $24$  & $50$   & $\infty$    & $50$   \\
CITESEER  & $175$    & $\infty$   & $175$  & $25$   & $\infty$    & $25$     \\
TEXAS     & $87$    & $0$   & $174$ & $37$  & $0$    & $74$   \\
CORNELL   & $100$    & $0$   & $200$  & $25$ & $0$   & $50$  \\
WISCONSIN & $25$    & $0$   & $50$  & $150$  & $\infty$    & $150$  \\
CHAMELEON & $50$    & $0$   & $100$  & $87$ & $0$    & $174$    \\
\midrule
ENZYMES   & $15$    & $0$  & $30$   & $5$    & $0$    & $10$       \\
IMDB      & $10$    & $0$     & $20$   & $10$    & $\infty$    & $10$ \\
MUTAG     & $20$    & $\infty$    & $20$  & $10$    & $0$    & $20$      \\
PROTEINS  & $5$    & $0$    & $10$   & $15$   & $0$    & $30$       \\
\bottomrule
\end{tabular}
    \end{small}
\end{sc}
\end{table}

\begin{table}[!t]
\centering
\caption{FoSR's best iteration count settings.}
\label{table:fosr_hyperparams}
\vskip 0.15in
\begin{sc}
\begin{small}
\begin{tabular}{lcc}
\toprule
Dataset & GCN & GIN \\
\midrule
CORA      & $150$ & $50$\\
CITESEER  & $100$ & $200$\\
TEXAS     & $50$  & $150$\\
CORNELL   & $125$ & $75$\\
WISCONSIN & $175$ & $25$\\
CHAMELEON & $50$  & $25$\\
\midrule
ENZYMES   & $40$ & $5$\\
IMDB      & $5$  & $20$\\
MUTAG     & $10$ & $20$\\
PROTEINS  & $30$ & $10$\\
\bottomrule
\end{tabular}
\end{small}
\end{sc}
\end{table}

\begin{table}[t!]
\centering
\caption{BORF's best hyper-parameter settings.} \label{table:borf_hyperparams}
\vskip 0.15in
\begin{sc}
    \begin{small}
        \begin{tabular}{lcccccccc}
\toprule
& \multicolumn{4}{c}{GCN}& \multicolumn{4}{c}{GIN}\\
\cmidrule(lr){2-5}
\cmidrule(lr){6-9}
Dataset &  $n$ &  $h$ &  $k$ & \# rewired &  $n$ &  $h$ &  $k$ & \# rewired \\

\midrule
CORA      & $3$    & $20$   & $10$   & $90$   & $3$    & $20$   & $30$   & $150$  \\
CITESEER  & $3$    & $20$   & $10$   & $90$   & $3$    & $10$   & $20$   & $90$   \\
TEXAS     & $3$    & $30$   & $10$   & $120$  & $1$    & $20$   & $10$   & $30$   \\
CORNELL   & $2$    & $20$   & $30$   & $100$  & $3$    & $10$   & $20$   & $90$   \\
WISCONSIN & $2$    & $30$   & $20$   & $100$  & $2$    & $50$   & $30$   & $160$  \\
CHAMELEON & $3$    & $20$   & $20$   & $120$  & $3$    & $30$   & $30$   & $180$  \\
\midrule
ENZYMES   & $1$    & $3$     & $2$    & $5$    & $3$    & $3$    & $1$    & $12$   \\
IMDB      & $1$    & $3$     & $0$    & $3$    & $1$    & $4$    & $2$    & $6$    \\
MUTAG     & $1$    & $20$    & $3$    & $23$   & $1$    & $3$    & $1$    & $4$    \\
PROTEINS  & $3$    & $4$     & $1$    & $15$   & $2$    & $4$    & $3$    & $14$   \\
\bottomrule
\end{tabular}
    \end{small}
\end{sc}
\end{table}

We report the best rewiring settings for every task and baseline GNN architecture. For SDRF, we set the temperature $\tau = \infty$ and only tuned the Ric upper bound $C^+$ and iteration count. For FoSR, we tuned the iteration count. For BORF, we tuned the number of batches $n$, number of edges added per batch $h$, and number of edges removed per batch $k$. The exact hyper-parameters for SDRF, FoSR and BORF are available in Table \ref{table:sdrf_hyperparams}, 
Table \ref{table:fosr_hyperparams}, and Table \ref{table:borf_hyperparams}, respectively. We also report the total amount of edges each method rewired, which equals the total number of added and removed connections for SDRF and BORF. The number of edges rewired by FoSR is the same as the iteration count.

 {For node classification tasks, each dataset is one big graph. The column titled ``\textsc{\# rewired}'' reports the total number of edges rewired in this graph. For graph classification tasks, each dataset is a collection of smaller graphs. The column titled ``\textsc{\# rewired}'' reports the number of edges rewired for each of these smaller graphs. We did not vary the number of edges rewired for each smaller graph in graph classification datasets.}

\subsection{Architecture and experiment settings}
\label{sec:gnn_opt_settings}
For graph and node classification, we utilized fixed model architectures with fixed numbers of GNN layers across all datasets. We used 3 GNN layers for node classification and 4 GNN layers for graph classification tasks. All the intermediate GNN layers (except for the input and output layers) have the same number of input and output dimensions specified by the hidden dimensions. After every GNN layer, we also added a drop-out layer with a fixed drop-out probability and an activation layer. The specific hidden dimensions, drop-out probabilities, and final activation layers for both node and graph classification tasks are specified in the architecture settings in Table \ref{table:exp_settings}.

\begin{table}[tb!]
\caption{Experiment settings for node and graph classification tasks (Note: The train fraction is with respect to the entire dataset while the validation fraction is with respect to the train set).}
\label{table:exp_settings}
\vskip 0.15in
\centering
\begin{sc}
\begin{small}
\begin{tabular}{lccccc}
\toprule
Task & Learning Rate & \#Trials/run & Stop patience & Train fraction & Validation fraction \\
\midrule
Node  & $0.001$ & $100$ & $100$ & $0.6$ & $0.2$ \\
Graph & $0.001$ & $100$ & $100$ & $0.8$ & $0.1$ \\

\bottomrule
\end{tabular}
\end{small}
\end{sc}
\end{table}

\begin{table}[tb!]
\caption{Architecture settings for node and graph classification tasks.}
\label{table:arch_settings}
\vskip 0.15in
\centering
\begin{sc}
\begin{small}
\begin{tabular}{lcccc}
\toprule
Task & Drop-out probability & \#GNN layers &
  Hidden Dimensions & Final activation \\
\midrule
Node  & $0.5$ & $3$ & $128$ & ReLU \\
Graph & $0.5$ & $4$ & $64$  & ReLU \\

\bottomrule
\end{tabular}
\end{small}
\end{sc}
\end{table}

For each graph and node classification experiment, we randomly split the dataset into train, validation and test sets $100$ times corresponding to $100$ trials. For each trial, the GNN model is trained on the train set using the Adam optimizer and validated using the validation set. The test accuracy corresponding to the best accuracy on the validation set is recorded as the test accuracy of the current trial. After all 100 trials are finished, the mean test accuracy and the $95\%$ confidence interval across all trials are computed and recorded in Tables \ref{table:node_classification_results} and \ref{table:graph_classification_results}. We also implemented a callback that stops the training process upon no improvement on the validation accuracy for $100$ epochs. The train and validation fractions used to split the dataset is specified in Table \ref{table:arch_settings}.

\section{Additional experiments and ablation studies}
\label{sec:additional_experiments}

\subsection{Ablation study - Effect of edge addition/removal}
\label{sec:edge_addition_removal}

In this section, we investigate the role of edge addition and removal in helping BORF improve GNN performance. We compare BORF's performance at high depths when using the best rewiring settings found in previous experiments against versions of BORF where it only removes edges, only adds edges (only alleviates over-squashing), or adds \& removes edges equally. Specifically, we consider the following cases
\begin{itemize}
    \item BORF with the best settings for 3 GNN layers on node classification datasets and 4 GNN layers for graph classification datasets, as reported in Table \ref{table:borf_hyperparams}.
    \item BORF removes the same number of edges $k$ as reported in Table \ref{table:borf_hyperparams} but adds $h = 0$ new edge.
    \item BORF adds the same number of edges $h$ as reported in Table \ref{table:borf_hyperparams} but removes $k = 0$ edge.
    \item BORF adds and removes an equal amount of edges $h=k$, taken to be the average of $h$ and $k$ as reported in Table \ref{table:borf_hyperparams}.
\end{itemize}

We use GCN as the baseline GNN and conduct our ablation study on all datasets used in previous experiments at depths 5, 7, and 9. Other experiment details are kept identical to previous experiments, as documented in Section \ref{sec:gnn_opt_settings}. We report the experiment results in tables \ref{table:ablation_study_edge_add_removal_node} and \ref{table:ablation_study_edge_add_removal_graph}. We note that the rewiring hyper-parameters were not tuned for higher depths. As such, the results in the columns ``\textsc{Best settings}'' are not the best performance achievable with BORF. With more depth-specific tuning, we expect to be able to obtain even better classification accuracy for both node classification and graph classification tasks.

\begin{table*}[t!]
\caption{Classification accuracies of GCN at depths 5, 7, and 9 with different BORF rewiring options on node classification datasets.}
\vspace{-0.5em}
\label{table:ablation_study_edge_add_removal_node}
\begin{center}
\resizebox{\linewidth}{!}{
\begin{sc}
\begin{tabular}{lcccccc}
\toprule
Dataset & \#layers & None & Best settings & Only add & Only remove & Remove + add equally \\
\midrule
\multirow{3}{*}{Cora} & 5 &
  $84.2 \pm 0.8$ &
  $85.1 \pm 1.3$ &
  $84.6 \pm 1.1$ &
  $84.9 \pm 0.9$ & 
  $85.0 \pm 0.7$ \\
& 7 &
  $82.1 \pm 1.1$ &
  $82.6 \pm 0.9$ &
  $81.8 \pm 1.2$ &
  $82.2 \pm 0.7$ & 
  $83.1 \pm 1.1$ \\
& 9 &
  $78.9 \pm 0.7$ &
  $78.4 \pm 1.2$ &
  $77.5 \pm 0.9$ &
  $78.2 \pm 1.4$ & 
  $77.9 \pm 1.2$ \\
\midrule

\multirow{3}{*}{Citeseer} & 5 &
  $69.2 \pm 1.1$ &
  $72.1 \pm 0.8$ &
  $70.5 \pm 1.4$ &
  $71.2 \pm 1.2$ & 
  $71.8 \pm 1.5$ \\
& 7 &
  $65.4 \pm 1.5$ &
  $68.3 \pm 0.9$ &
  $66.8 \pm 1.1$ &
  $67.5 \pm 0.9$ & 
  $68.1 \pm 1.2$ \\
& 9 &
  $61.7 \pm 0.9$ &
  $62.6 \pm 1.6$ &
  $61.5 \pm 1.2$ &
  $62.2 \pm 1.3$ & 
  $63.2 \pm 1.5$ \\
\midrule

\multirow{3}{*}{Texas} & 5 &
  $40.3 \pm 1.3$ &
  $43.3 \pm 0.7$ &
  $43.5 \pm 0.8$ &
  $41.1 \pm 1.0$ & 
  $42.8 \pm 0.9$ \\
& 7 &
  $37.2 \pm 1.1$ &
  $40.1 \pm 1.6$ &
  $41.4 \pm 1.2$ &
  $39.6 \pm 0.5$ & 
  $40.8 \pm 0.6$ \\
& 9 &
  $32.1 \pm 0.9$ &
  $34.4 \pm 1.2$ &
  $36.5 \pm 1.1$ &
  $35.1 \pm 0.7$ & 
  $35.8 \pm 0.6$ \\
\midrule

\multirow{3}{*}{Cornell} & 5 &
  $41.3 \pm 1.4$ &
  $45.5 \pm 1.1$ &
  $44.7 \pm 1.3$ &
  $46.4 \pm 1.2$ & 
  $45.9 \pm 1.2$ \\
& 7 &
  $39.5 \pm 1.7$ &
  $41.5 \pm 1.5$ &
  $42.8 \pm 1.4$ &
  $43.2 \pm 1.3$ & 
  $41.8 \pm 1.3$ \\
& 9 &
  $35.5 \pm 1.4$ &
  $40.9 \pm 1.3$ &
  $40.3 \pm 2.0$ &
  $41.9 \pm 1.6$ & 
  $39.9 \pm 1.6$ \\
  
\midrule
\multirow{3}{*}{Wisconsin} & 5 &
  $40.1 \pm 1.2$ &
  $47.2 \pm 0.6$ &
  $44.4 \pm 0.9$ &
  $45.2 \pm 1.1$ & 
  $45.1 \pm 0.7$ \\
& 7 &
  $35.5 \pm 1.1$ &
  $42.3 \pm 1.0$ &
  $39.7 \pm 0.9$ &
  $40.9 \pm 0.7$ & 
  $39.2 \pm 1.2$ \\
& 9 &
  $31.2 \pm 0.8$ &
  $36.4 \pm 0.6$ &
  $34.2 \pm 0.9$ &
  $35.8 \pm 1.2$ & 
  $34.7 \pm 1.1$ \\

\midrule
\multirow{3}{*}{Chameleon} & 5 &
  $57.2 \pm 0.4$ &
  $58.8 \pm 0.5$ &
  $58.1 \pm 0.7$ &
  $58.6 \pm 0.9$ & 
  $58.9 \pm 1.1$ \\
& 7 &
  $55.3 \pm 0.5$ &
  $56.1 \pm 0.8$ &
  $55.9 \pm 0.6$ &
  $56.3 \pm 0.4$ & 
  $55.8 \pm 0.9$ \\
& 9 &
  $51.1 \pm 0.7$ &
  $52.0 \pm 0.4$ &
  $51.8 \pm 0.6$ &
  $51.3 \pm 0.8$ & 
  $52.1 \pm 1.2$ \\

\bottomrule
\end{tabular}
\end{sc}}
\end{center}
\vskip -0.15in
\end{table*}

\begin{table*}[t!]
\caption{Classification accuracies of GCN at depths 5, 7, and 9 with different BORF rewiring options on graph classification datasets.}
\vspace{-0.5em}
\label{table:ablation_study_edge_add_removal_graph}
\begin{center}
\resizebox{\linewidth}{!}{
\begin{sc}
\begin{tabular}{ccccccc}
\toprule
Dataset & \#layers & None & Best settings & Only add & Only remove & Remove + add equally \\
\midrule
\multirow{3}{*}{Enzymes} & 5 &
  $24.2 \pm 1.3$ &
  $24.5 \pm 1.0$ &
  $24.8 \pm 1.6$ &
  $23.8 \pm 1.1$ & 
  $25.0 \pm 1.4$ \\
& 7 &
  $21.7 \pm 1.5$ &
  $22.3 \pm 1.2$ &
  $21.9 \pm 1.7$ &
  $21.1 \pm 1.2$ & 
  $22.1 \pm 1.3$ \\
& 9 &
  $20.8 \pm 0.9$ &
  $20.4 \pm 1.1$ &
  $20.8 \pm 1.2$ &
  $19.7 \pm 1.3$ & 
  $20.1 \pm 1.2$ \\
\midrule

\multirow{3}{*}{Imdb} & 5 &
  $47.6 \pm 0.8$ &
  $49.1 \pm 1.0$ &
  $48.8 \pm 1.1$ &
  $46.9 \pm 1.4$ & 
  $48.6 \pm 1.2$ \\
& 7 &
  $44.3 \pm 1.5$ &
  $45.4 \pm 1.2$ &
  $44.7 \pm 1.1$ &
  $43.7 \pm 1.3$ & 
  $44.2 \pm 1.4$ \\
& 9 &
  $39.2 \pm 1.0$ &
  $41.8 \pm 1.1$ &
  $40.7 \pm 1.6$ &
  $39.3 \pm 1.1$ & 
  $41.3 \pm 1.2$ \\
\midrule

\multirow{3}{*}{Mutag} & 5 &
  $67.7 \pm 1.6$ &
  $75.4 \pm 2.1$ &
  $76.1 \pm 2.2$ &
  $68.5 \pm 2.8$ & 
  $71.8 \pm 1.2$ \\
& 7 &
  $64.1 \pm 2.1$ &
  $72.1 \pm 1.3$ &
  $75.2 \pm 2.4$ &
  $65.1 \pm 1.5$ & 
  $66.2 \pm 1.9$ \\
& 9 &
  $63.1 \pm 1.2$ &
  $69.7 \pm 1.5$ &
  $70.4 \pm 1.7$ &
  $60.7 \pm 2.5$ & 
  $61.3 \pm 1.5$ \\
\midrule

\multirow{3}{*}{Proteins} & 5 &
  $69.2 \pm 0.8$ &
  $70.1 \pm 1.0$ &
  $69.5 \pm 1.1$ &
  $69.3 \pm 1.2$ & 
  $69.9 \pm 1.4$ \\
& 7 &
  $67.3 \pm 1.0$ &
  $68.1 \pm 0.9$ &
  $68.3 \pm 0.8$ &
  $67.5 \pm 1.1$ & 
  $67.9 \pm 1.2$ \\
& 9 &
  $64.5 \pm 1.1$ &
  $65.1 \pm 1.1$ &
  $64.9 \pm 1.2$ &
  $64.7 \pm 1.0$ & 
  $65.0 \pm 1.3$ \\
\bottomrule
\end{tabular}
\end{sc}}
\end{center}
\vskip -0.15in
\end{table*}

We observe that the un-rewired datasets record the worst performance in most scenarios considered. This indicates that GCN frequently suffers from both over-smoothing and over-squashing at high depths and that BORF helps the GNN achieve better performance even when it is restricted to only relieving over-smoothing or only relieving over-squashing.

\subsection{Experiment results on long-range graph benchmark}
We provide additional empirical results in this section to demonstrate the effectiveness of BORF on the long-range graph classification dataset Peptides-func introduced by \cite{dwivedi2023long}. In table \ref{table:lrgb_classification_results}, We compare three rewiring algorithms: SDRF, FoSR, and BORF by tuning the hyper-parameters of each method and report the mean average precision (mAP) of the best setting for each algorithm. For each hyper-parameter setting, we run our experiment with 4 random splits and 4 random seeds per split, similar to the setting in \cite{dwivedi2023long}. However, for our experiments, we utilized a lower hidden dimensions of 64 rather than 300. For both SDRF and FoSR, we performed hyper-parameters tuning by trying different numbers of iterations from 25 to 200 with an increment step of 25 to find the optimal number of rewiring iterations. For SDRF, we only tuned the hyper-parameter $C^+$ using values $0$ and $\infty$. For BORF, we set the range of rewiring batches from 1 to 4. For each number of rewiring batches, we tested for the following pairs of edge addition - removal settings: 30 - 20, 40 - 10. In table \ref{table:lrgb_classification_hyperparams}, we report the best settings of SDRF, FoSR, and BORF for both GCN and GIN layer types.

\begin{table*}[t!]
\caption{Classification accuracies of GCN and GIN with None, SDRF, FoSR, and BORF rewiring on the Peptides-func dataset. Best results are highlighted in bold.}
\label{table:lrgb_classification_results}
\begin{center}
\begin{small}
\begin{sc}
\begin{tabular}{lcccc}
\toprule
Layer type & None & SDRF & FoSR & BORF \\
\midrule
GCN &
  $40.1 \pm 2.1$ &
  $41.5 \pm 1.5$ &
  \boldsymbol{$44.2 \pm 2.1$} &
  $43.9 \pm 2.7$ \\
GIN & 
  $46.1 \pm 2.4$ &
  $46.3 \pm 1.7$ &
  $48.3 \pm 1.8$ &
  \boldsymbol{$50.2 \pm 1.7$} \\
\bottomrule
\end{tabular}
\end{sc}
\end{small}
\end{center}
\vskip -0.15in
\end{table*}

\begin{table*}[t!]
\caption{Hyper-parameter settings of SDRF, FoSR and BORF for GCN and GIN tested on the Peptides-func graph classification dataset.}
\label{table:lrgb_classification_hyperparams}
\begin{center}
\begin{small}
\begin{sc}
\begin{tabular}{lccc}
\toprule
Layer type & SDRF & FoSR & BORF \\
\midrule
GCN &
  100 iterations - $C^+ = 0$ &
  25 iterations &
  4 batches - add 30 - remove 20 \\
GIN & 
  50 iterations - $C^+ = 0$ &
  150 iterations &
  2 batches - add 40 - remove 10
  \\
\bottomrule
\end{tabular}
\end{sc}
\end{small}
\end{center}
\vskip -0.15in
\end{table*}

We observe that on GCN, BORF’s performance is comparable with that of FoSR. Both of these rewiring algorithms significantly improve the model performance compared to the baseline and SDRF. On GIN, BORF is the best performer overall.

\subsection{Graph topology changes by different rewiring algorithms}
In this section, we provide empirical data comparing the change in graph topology enacted by the rewiring algorithms SDRF, FoSR, and BORF. Similar to \cite{topping2022}, for each dataset, we record the node degrees' base 2 logarithm distribution before and after applying BORF, SDRF and FoSR rewiring settings tuned for GCN as documented in Section \ref{sec:hyperparams}. We visualize the difference by utilizing the kernel density functions. The $L_1$ Wasserstein distance between the kernel density functions of these rewiring methods and the original node degree density is also calculated to measure the extent of topological change after rewiring.

\newcolumntype{C}{>{\centering\arraybackslash}X}
\begin{figure}[!ht]
\label{fig:topology_change}
\begin{tabularx}{\linewidth}{CCC}
    \begin{subfigure}{\linewidth}
        \includegraphics[width=\linewidth]{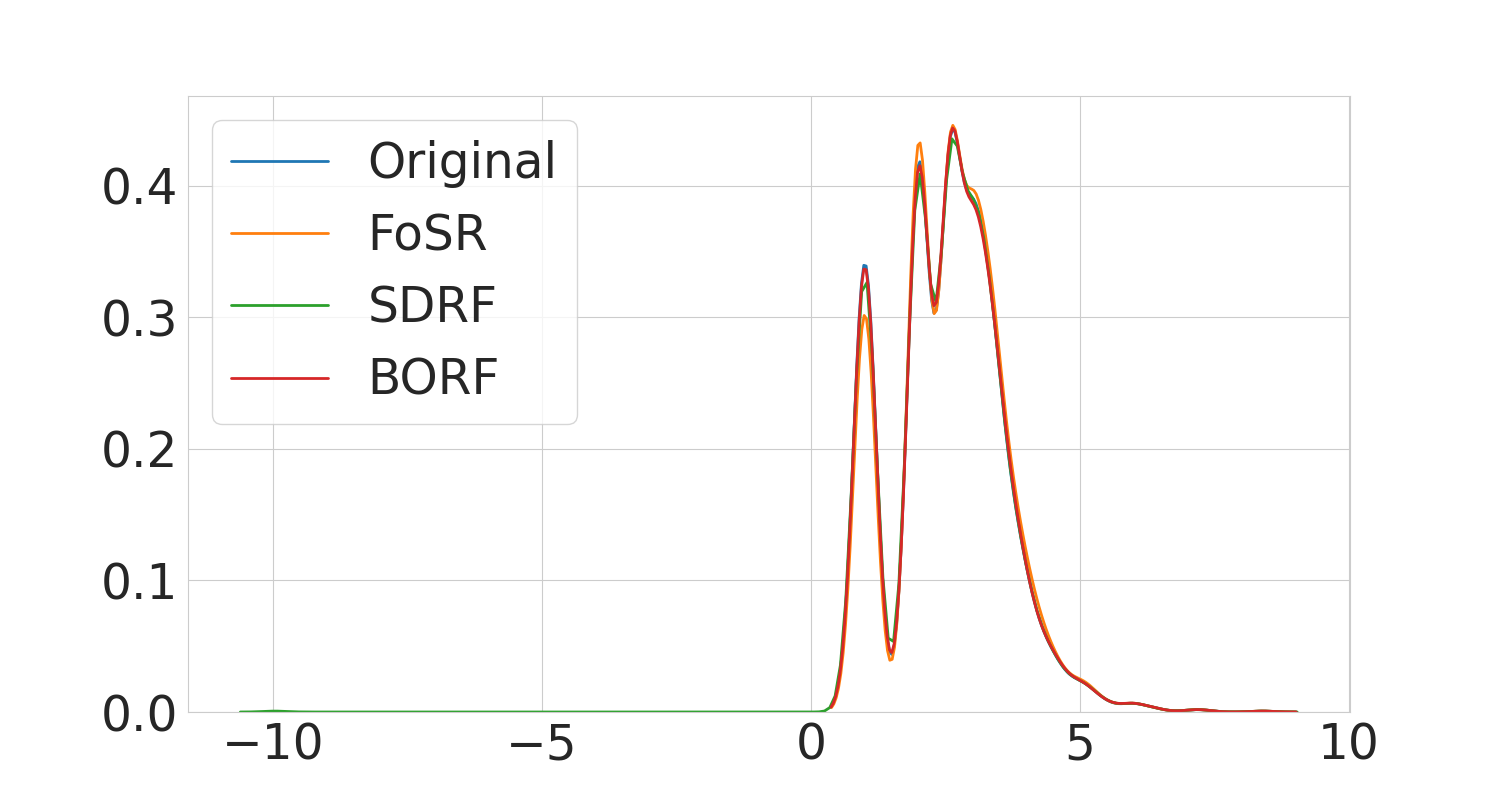}
    \caption{Cora\\$W_1(\text{Original, SDRF})=0.00115$\\$W_1(\text{Original, FoSR})=0.00372$\\$W_1(\text{Original, BORF})=0.00082$}
    \end{subfigure}

    \begin{subfigure}{\linewidth}
        \includegraphics[width=\linewidth]{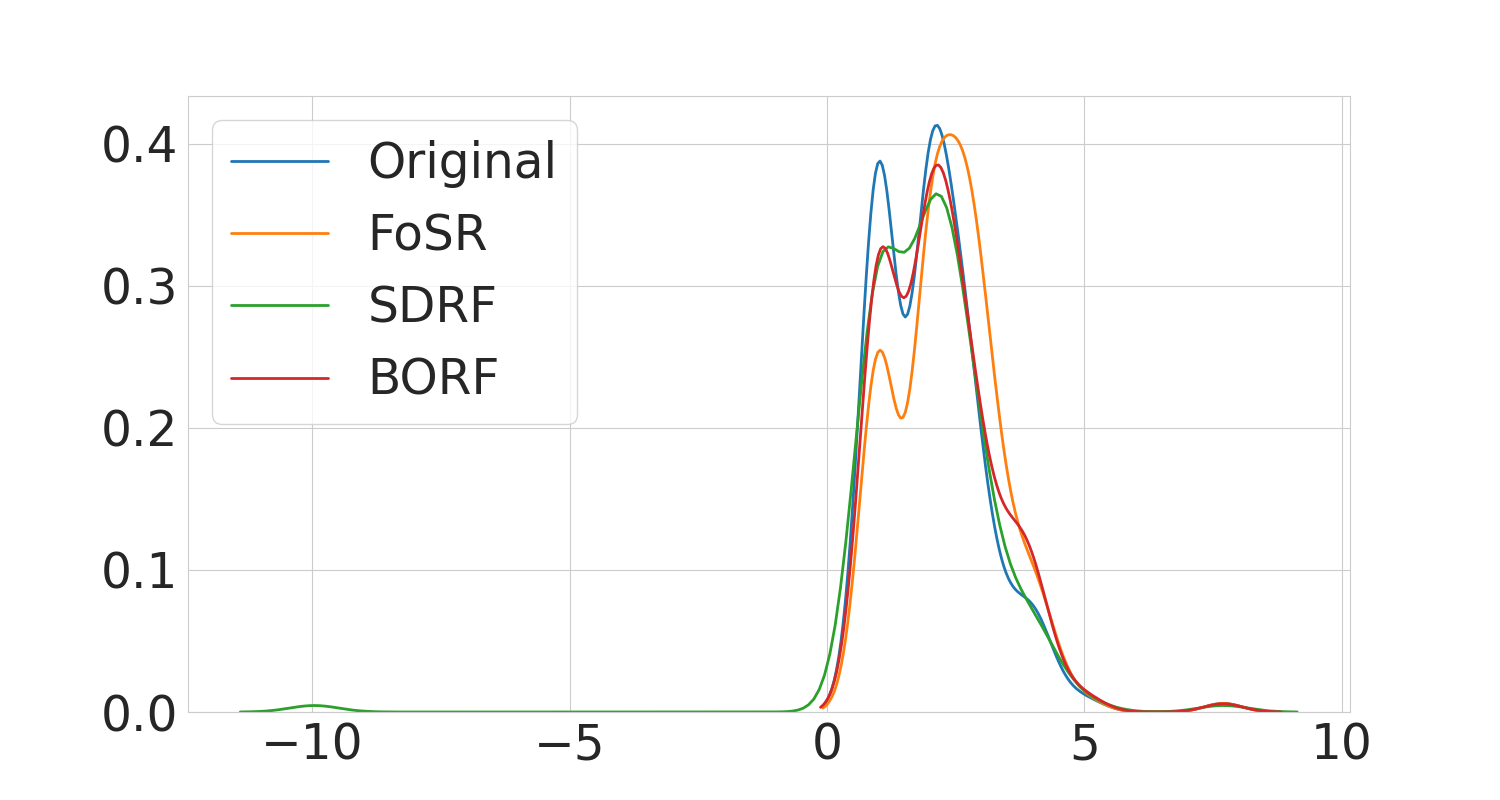}
    \caption{Texas\\$W_1(\text{Original, SDRF})=0.00831$\\$W_1(\text{Original, FoSR})=0.01024$\\$W_1(\text{Original, BORF})=0.00971$}
    \end{subfigure} &

    \begin{subfigure}{\linewidth}
        \includegraphics[width=\linewidth]{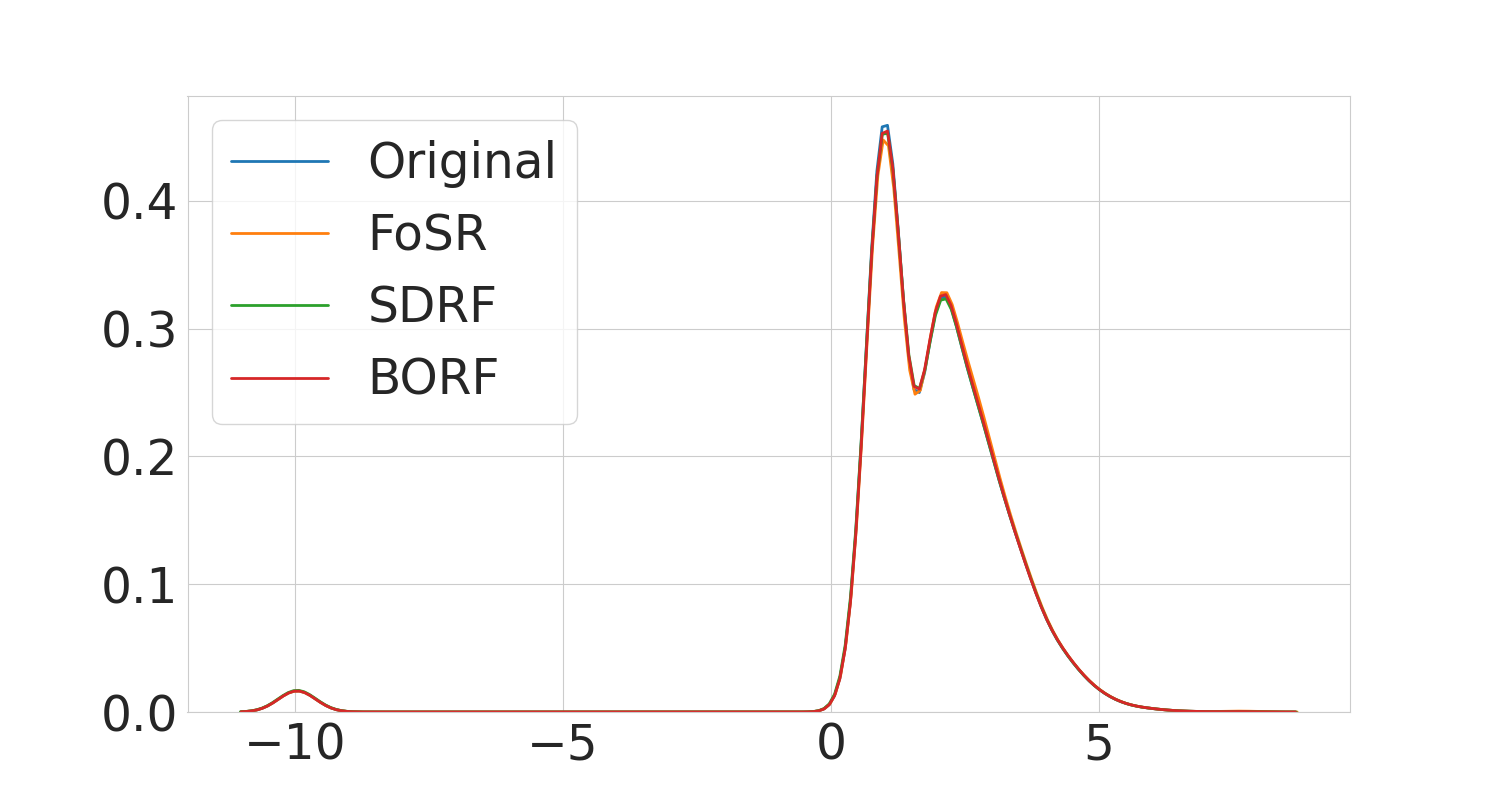}
    \caption{Citeseer\\$W_1(\text{Original, SDRF})=0.00056$\\$W_1(\text{Original, FoSR})=0.00184$\\$W_1(\text{Original, BORF})=0.00072$}

    \end{subfigure}
    \begin{subfigure}{\linewidth}
        \includegraphics[width=\linewidth]{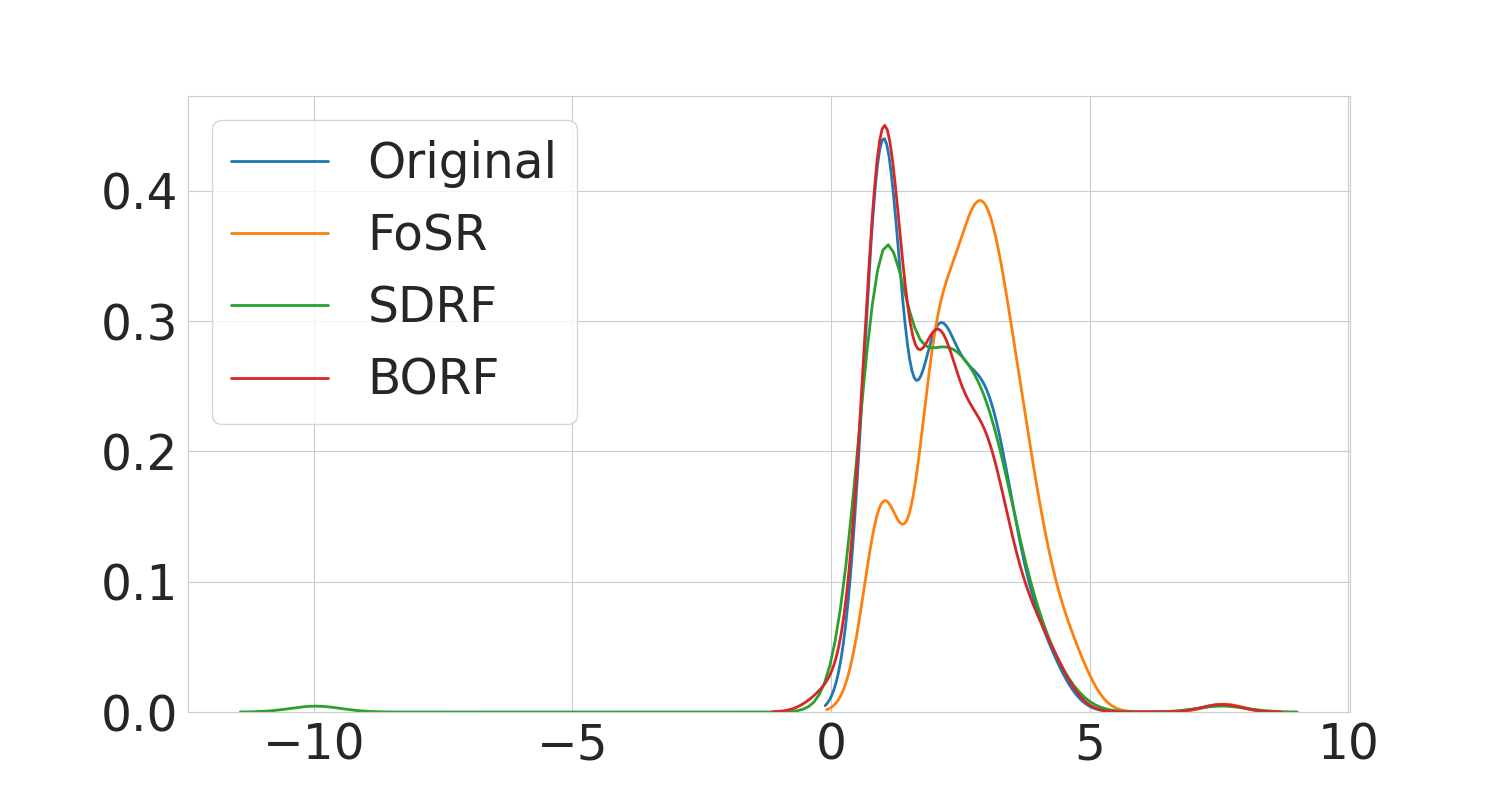}
    \caption{Cornell\\$W_1(\text{Original, SDRF})=0.00857$\\$W_1(\text{Original, FoSR})=0.01534$\\$W_1(\text{Original, BORF})=0.00641$}
    \end{subfigure} &

    \begin{subfigure}{\linewidth}
        \includegraphics[width=\linewidth]{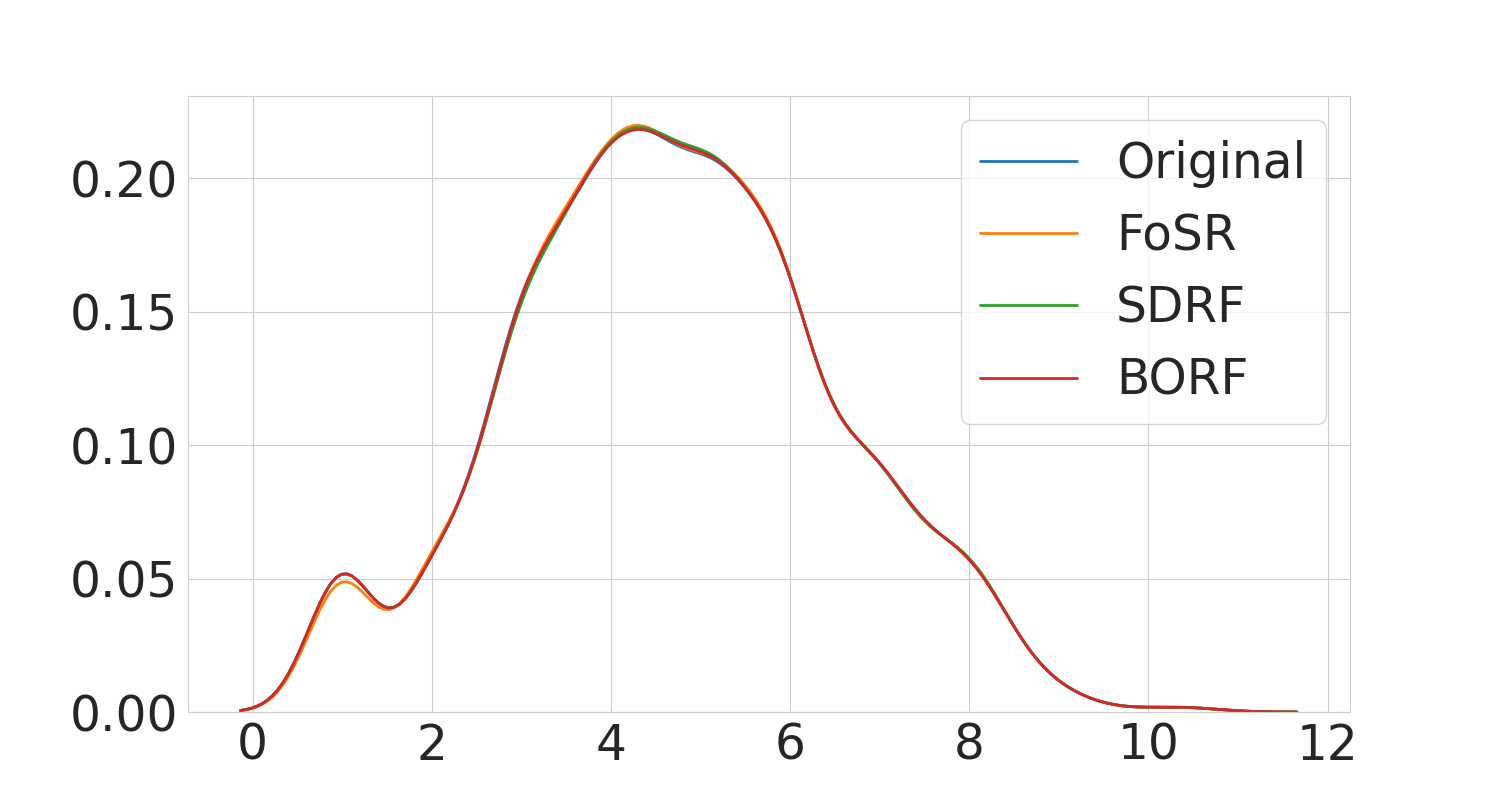}
    \caption{Chameleon\\$W_1(\text{Original, SDRF})=0.00043$\\$W_1(\text{Original, FoSR})=0.00069$\\$W_1(\text{Original, BORF})=0.00026$}
    \end{subfigure}

    \begin{subfigure}{\linewidth}
        \includegraphics[width=\linewidth]{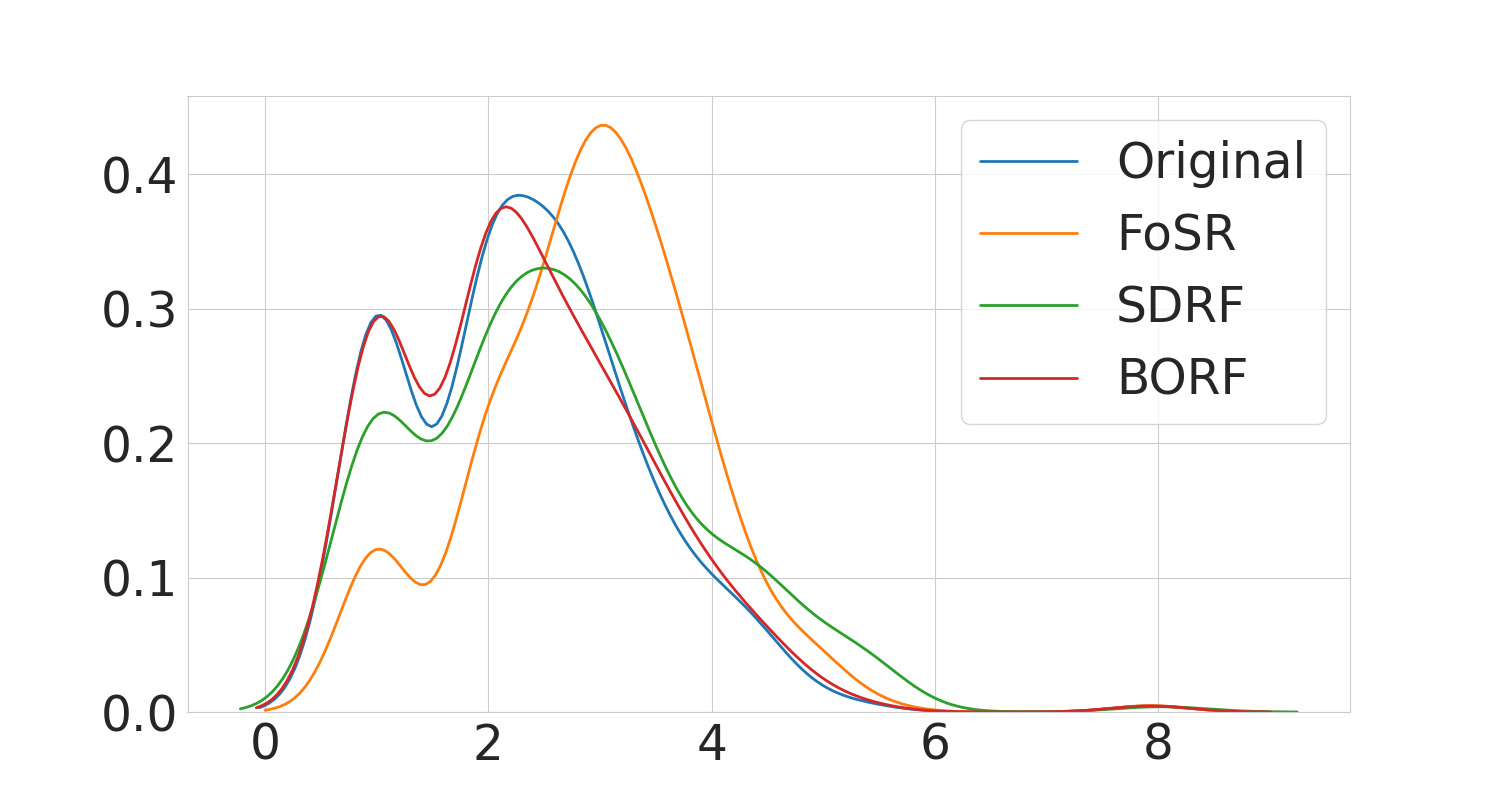}
        \caption{Wisconsin\\$W_1(\text{Original, SDRF})=0.00836$\\$W_1(\text{Original, FoSR})=0.01723$\\$W_1(\text{Original, BORF})=0.00451$}
    \end{subfigure}
\end{tabularx}
\caption{Kernel density functions of graph degrees' base 2 logarithm before and after rewiring using BORF, SDRF and FoSR.}
\end{figure}

Our data shows that both BORF and SDRF change the graph topology only minimally according to this metric, while rewiring with FoSR can have a much more drastic effect.




\section{Dataset Statistics}
\label{sec:dataset_statistics}
We provide a summary of statistics of all datasets used in Table \ref{table:node_classification_dataset_statistics} and Table \ref{table:graph_classification_dataset_statistics}. We also report the mean and standard deviation of the Ollivier Ricci curvature for each dataset. On node classification tasks, this is exactly the statistics of the set of edge curvature values. On graph classification tasks, this is the statistics of the mean curvature value of all graphs within the dataset.
\begin{table}[t!]
\centering
\caption{Statistics of node classification datasets.}
\label{table:node_classification_dataset_statistics}
\vskip 0.15in
\begin{sc}
\begin{small}
\begin{tabular}{lcccccc}
\toprule
\multicolumn{1}{c}{\textbf{}} & Cornell & Texas & Wisconsin & Cora & Citeseer & Chameleon \\
\midrule 
\#Nodes    & 140  & 135  & 184  & 2485  & 2120  & 832      \\
\#Edges    & 219  & 251  & 362  & 5069  & 3679  & 12355    \\
\#Features & 1703 & 1703 & 1703 & 1433  & 3703  & 2323     \\
\#Classes  & 5    & 5    & 5    & 7     & 6     & 5        \\
Directed   & TRUE & TRUE & TRUE & FALSE & FALSE & TRUE       \\
{ORC Mean} & -0.39 & -0.24 & -0.59 & -0.19 & -0.31 & 0.64 \\
{ORC STD} & 0.52 & 0.45 & 0.71 & 0.68 & 0.78 & 0.58 \\
\bottomrule
\end{tabular}
\end{small}
\end{sc}
\end{table}

\begin{table}[t!]
\centering
\caption{Statistics of graph classification datasets.}
\label{table:graph_classification_dataset_statistics}
\vskip 0.15in
\begin{sc}
\begin{small}
\begin{tabular}{lcccc}
\toprule
\multicolumn{1}{c}{\textbf{}} & Enzymes & Imdb      & Mutag   & Proteins  \\
\midrule
\#Graphs                      & 600     & 1000      & 188     & 1113      \\
\#Nodes                       & 2-126   & 12-136    & 10 - 28 & 4-620     \\
\#Edges                       & 2 - 298 & 52 - 2498 & 20 - 66 & 10 - 2098 \\
Avg \#Nodes                   & 32.63   & 19.77     & 17.93   & 39.06     \\
Avg \#Edges                   & 124.27  & 193.062   & 39.58   & 145.63    \\
\#Classes                     & 6       & 2         & 2       & 2         \\
Directed                      & FALSE   & FALSE     & FALSE   & FALSE     \\
{ORC Mean}                    & 0.13             & 0.58          & -0.27                              & 0.17              \\
{ORC STD}                     & 0.15             & 0.19          & 0.05                               & 0.20  \\   
\bottomrule
\end{tabular}
\end{small}
\end{sc}
\end{table}

\section{Hardware Specifications and Libraries}
\label{sec:hardware_library}

All experiments were implemented in Python using PyTorch \cite{Paszke_PyTorch_An_Imperative_2019}, Numpy \cite{2020NumPy-Array}, PyG (PyTorch Geometric) \cite{Fey_Fast_Graph_Representation_2019}, POT (Python Optimal Transport) \cite{flamary2021pot} with figures created using TikZ \cite{tantau2023tikz}. PyTorch, PyG and NumPy are made available under the BSD license, POT under MIT license, and TikZ under the GNU General Public license.

We conducted our experiments on two local servers with the specifications laid out in Table \ref{table:hardware_specifications}.
\begin{table}[t!]
\centering
\caption{Server specifications for conducting all experiments.}\label{table:hardware_specifications}
\vskip 0.15in
\begin{sc}
\begin{small}
\begin{tabular}{lll}

\toprule
\multicolumn{1}{c}{{Server ID}} & {Components} & {Specifications}           \\
\midrule
\multirow{5}{*}{\textbf{1}}         & Architecture        & X86\_64                            \\
                                    & OS                  & Ubuntu 20.04.5 LTS x86\_64         \\
                                    & CPU                 & Intel i7-10700KF (16) @ 5.100GHz  \\
                                    & GPU                 & NVIDIA GeForce RTX 2080 Ti Rev. A \\
                                    & RAM                 & 12Gb                              \\
\midrule
\multirow{5}{*}{\textbf{2}}         & Architecture        & X86\_64                            \\
                                    & OS                  & Ubuntu 20.04.5 LTS x86\_64         \\
                                    & CPU                 & AMD EPYC 7742 64-core             \\
                                    & GPU                 & NVIDIA A100 Tensor Core           \\
                                    & RAM                 & 40Gb\\
\bottomrule
\end{tabular}
\end{small}
\end{sc}
\end{table}



\end{document}